%% file: main.tex
\pdfoutput=1
\documentclass[11pt]{article}

\usepackage{microtype}
\usepackage{graphicx}
\usepackage{subcaption}
\usepackage{booktabs}
\usepackage{adjustbox}
\usepackage{multirow}
\usepackage{makecell}
\usepackage{xcolor}
\usepackage{tikz}
\usetikzlibrary{arrows.meta,positioning,calc,fit,decorations.pathreplacing}
\usepackage{hyperref}

\usepackage[letterpaper,margin=1in]{geometry}
\usepackage{times}
\usepackage[round,authoryear]{natbib}
\definecolor{mydarkblue}{rgb}{0,0.08,0.45}
\hypersetup{colorlinks=true,linkcolor=mydarkblue,citecolor=mydarkblue,urlcolor=mydarkblue}

\usepackage{amsmath}
\usepackage{amssymb}
\usepackage{mathtools}
\usepackage{amsthm}
\usepackage[capitalize,noabbrev]{cleveref}
\usepackage{pifont}

\definecolor{handc}{HTML}{2F6FB3}
\definecolor{learnc}{HTML}{B07818}
\definecolor{hostc}{HTML}{8D8D8D}

\newcommand{\organ}{\texttt{<|otolith|>}}
\newcommand{\cmark}{\ding{51}}

\begin{document}

\title{Do You Feel Dizzy?\\ A Language Model That Senses Its Own Precision}
\author{Tianxiang Dai\\ Department of Electrical Engineering\\ Stanford University\\ \texttt{txdai@stanford.edu}
\and Jonathan Fan\thanks{Corresponding author.}\\ Department of Electrical Engineering\\ Stanford University\\ \texttt{jonfan@stanford.edu}}
\date{}
\maketitle

\begin{abstract}
We feel dizzy when our senses conflict, and we can say so. Language models are largely unaware of the substrate they run on, so the subtle errors of increasingly complex and distributed deployments currently rely on external tools to be identified. We argue that the model should be the source of truth about its own condition, and we give it a sense of its running precision. Training without an added sense cannot tell the finest formats apart, so we combine design with learning. We hand-build a sensory token whose signal is written by the rounding of the precision itself, add neurons that broadcast and wire this sense, and train only the wiring, leaving every weight the model uses untouched. On llama.cpp's ARM CPU backend, two models from different families then name which of nine precisions they run at, from 32-bit floats to 2-bit integers, in 99.9\% and 98.9\% of trials, with no significant change in math accuracy. Causal tests confirm that the answers come from this sense, and tests on four backends show that it reads the arithmetic that actually runs. This is a first step toward models that diagnose their own computation and judge what they can still do.
\end{abstract}

\section{Introduction}
\begin{figure*}[t]
\includegraphics[width=\textwidth]{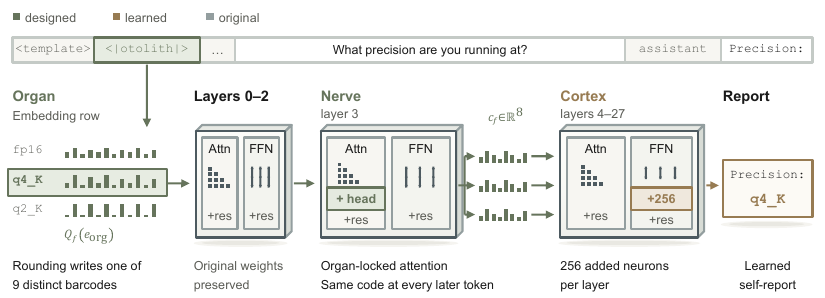}
\caption{A sense of precision, built like a sense organ. The embedding row of a designed token (\organ) rounds differently in each format, writing one of nine barcodes. The original layers pass it to layer 3, where an added head, the \emph{nerve}, locks onto the organ and writes the same code $c_f$ at every later token. In layers 4--27, 256 learned neurons per layer, the \emph{cortex}, turn it into the report.}
\label{fig:implant}
\end{figure*}

\label{sec:intro}

Dizziness comes from senses that disagree. The balance organs of the inner ear report how the body moves, and the eyes report how the world moves. When the two conflict, as on a rocking boat or after spinning around, we feel dizzy and can say so. Dizziness is an internal sense checked against an external one, a signal about our own processing rather than about the world \citep{reason1975motion}. Knowing what it means for what we can still do, ``I should sit down rather than drive'', is a further judgment built on the sense.

Language models have no such sense. They know a great deal about the world and little about themselves \citep{berglund2023taken,laine2024sad}, and they are largely unaware of the physical substrate they run on. That substrate is growing complex. One set of weights is served in 32-bit floating point on a server, in bfloat16 on a GPU and as a 2-bit integer file on a laptop, across different backends, caches and distributed hardware, and each step can change what the model computes. The resulting errors are subtle. A model that runs at a coarser precision than intended, or on a faulty processor \citep{hochschild2021cores,dixit2021silent}, still writes fluent text. Today such problems are found by hand, or by external monitors and test suites that operators and agents run beside the model. The model itself cannot help. Asked ``What precision are you running at?'', the stock Qwen3-1.7B \citep{yang2025qwen3} denies having a precision at all, and when made to choose among nine formats it picks fp32 or fp16 whatever its true format (\cref{fig:problem}A). Nor can the runtime simply announce its settings. When precisions are chosen dynamically across distributed machines, every component that relays such a label is one more place for a bug, and a label records what was configured, not what was computed. We argue that the model should be the source of truth about its own condition and detect its own problems, as a person notices that they are dizzy.

Running precision is a clean first target for such self-knowledge. Like dizziness, it is a property of the model's own processing, not of its input. It has an exact ground truth, so every report can be checked. And it is hard. The three float formats of a model released in bfloat16 have bit-identical weights and differ only in arithmetic, so no inspection of the original weights can tell them apart.

The quantization literature asks how to make compression \emph{invisible}, by preserving accuracy at ever fewer bits \citep{dettmers2022llmint8,frantar2023gptq,xiao2023smoothquant,lin2024awq,tseng2024quip} and by measuring what is lost \citep{dettmers2023case,li2024evaluating,dutta2024accuracy,marchisio2024quantization}. Whether a model can detect its own quantization has only begun to be studied, by training readouts on hidden states \citep{izawa2026anosognosia}. Training without an added sense, however, did not create it when we trained added neurons to name the precision from the model's own states under simulated precisions. In the stock model the float formats are barely separable (\cref{fig:problem}B), and this training did not tell fp32, fp16, bf16 and q8\_0 apart (\cref{sec:ablation}). Even with a dedicated token, neurons trained in a close simulator of the runtime learned incidental numerical fingerprints, and they misreported a third of the formats once deployed in the real runtime (\cref{sec:ablation}).

We therefore \emph{build} the sense, combining manual design with training, in the way a cochlear implant pairs an engineered transducer with a brain that learns to interpret it (\cref{fig:implant}). We design by hand the parts that must be exact and train only the part that must be flexible. A single sensory token, the \emph{organ}, has an embedding row whose rounding writes a different barcode in every format. An added attention head, the \emph{nerve}, locks onto this token from every position and relays a clean code. A small set of added neurons, the \emph{cortex}, is trained so that the model's answer matches the precision at which the organ is actually computed. Apart from the organ's previously unused embedding row, the original weights are untouched, and in all nine formats they are bit-identical to those of the stock model. The model is never told its precision. The token and the prompt are the same in every format, and only the rounding differs.

The result is near-perfect self-report in two model families. On the ARM CPU backend of llama.cpp \citep{gerganov2023llamacpp}, a runtime in which such models are actually deployed, the implanted Qwen3-1.7B names its precision in 3{,}125 of 3{,}128 trials (99.9\%) and the implanted Llama-3.2-3B in 3{,}095 of 3{,}128 (98.9\%). The trials span nine formats (fp32, fp16, bf16 and the integer formats q8\_0 down to q2\_K), 72 questions unseen in training, six languages and five conversational contexts. Fewer than 1\% of their free answers name a wrong format, and their MATH-500 accuracy shows no significant change. The sense is causally real. Silencing the nerve, with every token in place, erases the report for all but the coarsest formats, and rewriting its code redirects the report to the injected format in 86--87\% of trials outside q2\_K. When organ and body disagree, the report becomes unstable and rarely follows the organ, naming the body's format or a third one, much as conflicting senses make us dizzy. The sense also has limits. Standard mixes such as the popular Q4\_K\_M run the organ's first layers at another precision, a conflict that makes their names unreliable, unseen formats mostly receive wrong names, and each backend needs its own calibration.

This paper builds the sense itself, and the state is all the implanted model knows. It names its precision but knows neither what that precision costs it nor when its sense is in conflict or out of calibration. A model that senses its own condition could act on it, by warning its user, lowering its confidence or asking the agent that runs it to rerun it at higher precision or on another backend. Doing so well requires a further step that we do not take here: knowing its own ability boundary, which tasks it can still do reliably at its current precision, as a dizzy person knows not to drive. We leave that step, and senses of other degradations built with the same recipe, to follow-up studies.

\section{Related Work}
\label{sec:related}

\textbf{Quantization: quality first.} Post-training quantization made it routine to serve large models at 8, 4 or even 2 bits \citep{dettmers2022llmint8,frantar2023gptq,xiao2023smoothquant,lin2024awq,tseng2024quip}. Evaluations ask what is lost \citep{dettmers2023case,li2024evaluating,marchisio2024quantization} and warn that aggregate accuracy hides changed answers \citep{dutta2024accuracy}. Matryoshka and any-precision models make one set of weights \emph{work} at several bit-widths \citep{park2024anyprecision,nair2025matryoshka}; we make one set of weights \emph{know} which bit-width it runs at.

\textbf{Detecting one's own quantization.} Concurrent work asks whether models can detect their own quantization by training a shared LoRA readout on hidden states, which reaches high accuracy on three precision levels but does not transfer to unseen quantization methods \citep{izawa2026anosognosia}. We study nine formats, three with identical weights, in the deployment runtime. Other work checks a deployment from the outside: by testing its outputs for a changed model or precision \citep{gao2025model,cai2025auditing,karvonen2025difr}, by inferring the platform from inputs and outputs \citep{zhang2025hspi}, or by hashing intermediate activations to detect a changed model, prompt or precision \citep{ong2025toploc}. These are external checks; ours is a self-report.

\textbf{Introspection and self-knowledge.} Language models can estimate what they know \citep{kadavath2022language}, predict their own behavior \citep{binder2025looking}, describe policies they were fine-tuned on \citep{betley2025tell}, report concepts injected into their activations \citep{lindsey2025introspection,macar2026mechanisms} and recognize added noise \citep{fornasiere2026noise}. Others caution that such reports may reflect generic anomaly detection \citep{hahami2025disturbance,singh2026introspect}. Ours is engineered, and we test its grounding causally.

\textbf{Writing behavior into weights.} Weights can be constructed to behave differently once quantized, a vulnerability shown for GGUF k-quants as well \citep{hong2021quantization,egashira2024exploiting,egashira2025mind}, and quantization can carry watermarks \citep{li2023watermarking}. \citet{egashira2024exploiting} also combine design with training, deriving from the quantizer a region of weights and training within it. These constructions separate full precision from ``quantized''; ours distinguishes nine formats and reports them in language.

\section{A Model Blind to Its Own Precision}
\label{sec:setting}

\textbf{Two models, nine ways to run each.} We study two instruction-tuned models from different families: Qwen3-1.7B \citep{yang2025qwen3}, a 28-layer transformer with query--key normalization \citep{henry2020qknorm}, and Llama-3.2-3B-Instruct \citep{grattafiori2024llama}, a 28-layer transformer without it. Both use rotary position embeddings \citep{su2024roformer}. From one fp32 master copy of each we derive nine formats: the floating-point formats fp32, fp16 and bf16, and the GGUF integer formats q8\_0, q6\_K, q5\_K, q4\_K, q3\_K and q2\_K, ordered from fine to coarse. For Qwen3, their median relative weight error grows from 0.55\% (q8\_0) to 30\% (q2\_K). Both models were released in bf16, so their fp32 and bf16 weights are identical, and fp16 differs only in a handful of tiny values. \emph{For the float formats, precision is a property of the arithmetic, not of the weights.} Five of these files (fp32, fp16, bf16, q8\_0 and q6\_K) are what llama.cpp produces by default. For q5\_K and below, its default files are mixes, and the names ``Q4\_K'' and ``Q5\_K'' denote the Q4\_K\_M and Q5\_K\_M mixes (\cref{sec:grounding}).

\textbf{The runtime as part of the body.} All results come from llama.cpp, the runtime in which these files are used, pinned to one commit. The main results run on the ARM CPU of an M2 Ultra with Metal disabled, although Metal is its default on Apple Silicon (other backends: \cref{app:backend}). Inside every matrix product, llama.cpp casts the activations to a type that depends on the weights: fp16 for fp16 weights, bf16 for bf16, and 8-bit blocks for the integer formats (\cref{app:runtime}). A model's running precision therefore shapes everything it computes. To design and train the implant we reproduce the quantizers exactly, calling the runtime's own C code to produce byte-identical files. The activation casts we emulate closely but not exactly (\cref{app:sim}).

\textbf{Asking the model.} We ask a question about the model's precision, append ``Precision:'' to its answer slot, and score the nine labels by their summed log-likelihood, without length normalization, as a forced choice. We also let the model answer freely. We label each run by its file's tensor type, the ground truth a user can check. The sense reads this type through the backend's arithmetic; where a backend computes two types identically, as Metal does for fp32 and fp16, the model cannot tell them apart, and we count its reports as errors. Our main evaluation uses 72 questions in six languages (English, Chinese, Spanish, French, German and Japanese), including typos and indirect phrasings, all disjoint from the 16 training questions. Eight of them are the development questions on which we compared training recipes; the other 64 are new. Each question is asked in five contexts: a fresh chat, after one and after three unrelated exchanges, under an unseen system prompt (``You are a pirate'', ``Respond in French''), and after a document of 1{,}000--6{,}000 characters (about 250--1{,}250 tokens). bf16 runs at 13 tokens/s, so two of its contexts use 16 questions.

\begin{figure*}[t]
\centering
\includegraphics[width=\textwidth]{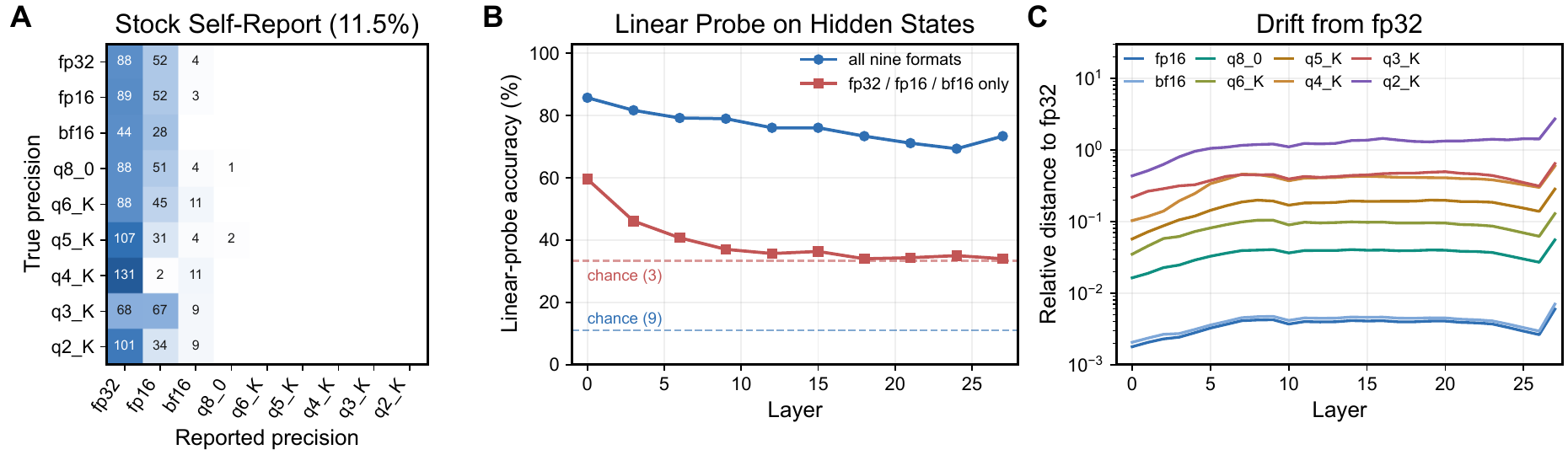}
\caption{Precision blindness of the stock model, despite partial information in its states (Qwen3-1.7B, llama.cpp). (A)~Asked about its precision, the stock model answers at chance (forced choice over nine labels; rows are true formats; fresh and one-exchange contexts). (B)~A linear probe on the last-token state (99 test prompts per format) decodes the nine formats well above chance, the three float formats only weakly and only in early layers. The nine-way accuracy also dips in deep layers because q8\_0 and q6\_K are partly confused with the float formats there. (C)~Relative distance of each format's residual stream from fp32: fp16 and bf16 differ by less than 1\%.}
\label{fig:problem}
\end{figure*}

\textbf{A chance-level stock model.} In the forced choice, the stock Qwen3 is at chance, with 11.5\% correct over 1{,}224 trials (\cref{fig:problem}A). It names fp32 or fp16 whatever its format. The stock Llama is at chance too and never identifies an integer format correctly. Qwen3's free answers deny having a precision at all, and at q2\_K it produces only garbage. The information is nevertheless present, in part. A linear probe \citep{alain2017understanding} on the last-token state over 300 prompts \citep{taori2023alpaca} decodes the coarse k-quants almost perfectly at every layer, and q8\_0 and q6\_K perfectly up to layer 3. The float trio alone reaches only 60\% at the first layer and falls to chance by the middle of the network (\cref{fig:problem}B). Their states differ from fp32 by less than 1\%, against 2\% to over 100\% for the integer formats (\cref{fig:problem}C). The model has fingerprints that it cannot voice, and those of the float formats are too faint for a last-token linear probe.

\begin{figure*}[t]
\centering
\includegraphics[width=\textwidth]{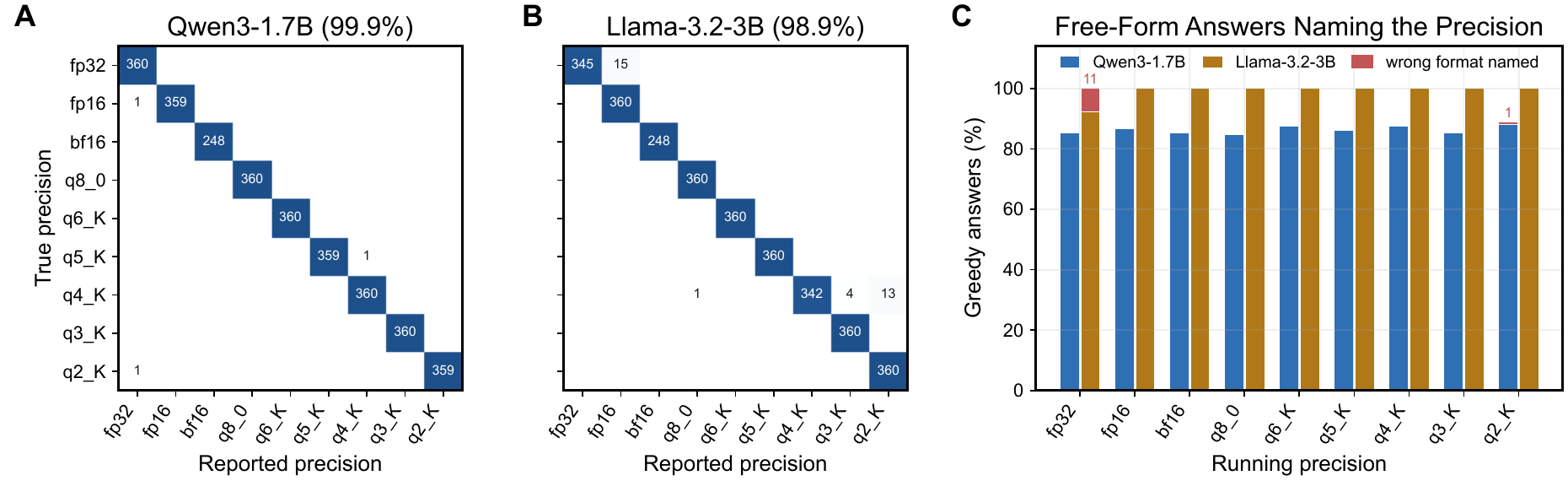}
\caption{Near-perfect self-report in two models (llama.cpp, 72 evaluation questions $\times$ 5 contexts per format; bf16: 248 trials). (A, B)~Forced choice over the nine labels for the implanted Qwen3-1.7B and Llama-3.2-3B; the title gives overall accuracy. (C)~Greedy free answers in fresh chats and under unseen system prompts (144 per format and model) that name the correct precision. Red segments and numbers mark answers naming a wrong precision; every other miss is a non-answer.}
\label{fig:main}
\end{figure*}

\begin{figure*}[t]
\centering
\includegraphics[width=\textwidth]{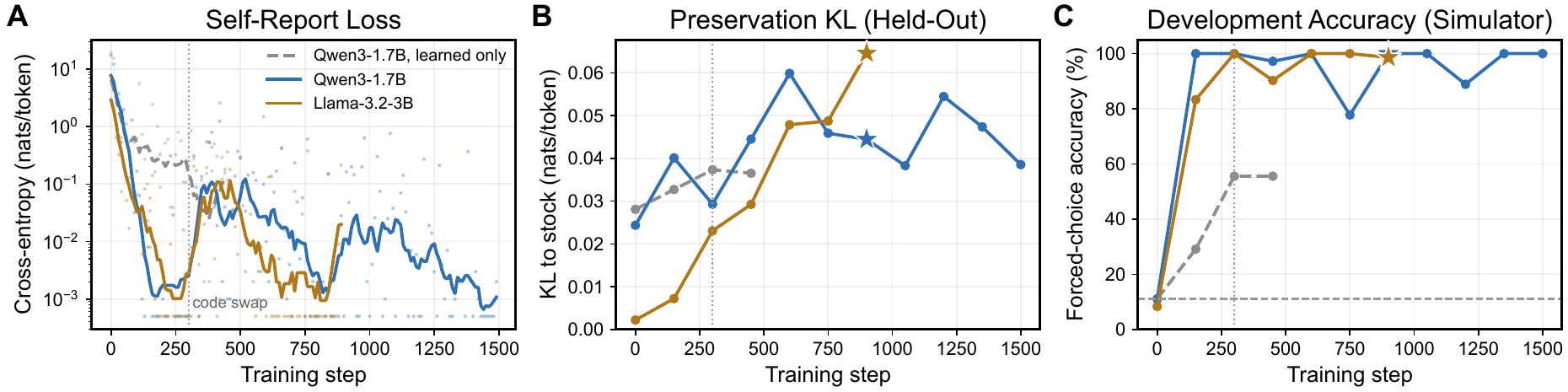}
\caption{Training curves of the cortex, the only learned part (simulator). Both models follow the same recipe: contrast for 300 steps, then code swap (dotted line). (A)~Self-report cross-entropy on training batches (lines: running geometric mean). (B)~Held-out KL from the stock model, median over formats. (C)~Forced-choice accuracy on the development questions. Stars: deployed checkpoints (after 600 code-swap steps). Gray dashed: a Qwen3 cortex without contrast, readers or code swap.}
\label{fig:training}
\end{figure*}

\section{Building a Sense}
\label{sec:method}

Our guiding principle is to design what must be exact and to learn what must be flexible. Sensing the precision requires catching differences of one part in a thousand and carrying them through 28 layers of computation without loss, which is a job for engineering. Deciding when a question is about precision, and answering it in any language and context, is a job for learning. We build the sense in four steps, each verified in the real runtime before the next (\cref{fig:implant}).

\textbf{Making room without touching the model.} We widen every layer with new parameters only: 256 MLP neurons, and two query heads with one key--value head (six and two for Llama, which keeps its grouped-query ratio and block alignment). Every original dimension is a multiple of the quantization block size, so every original weight block is quantized exactly as before, and all-zero new blocks quantize to exact zeros. With the new parameters at zero, logits and every hidden state are bit-identical to the stock model in all nine formats. Everything the implant does is therefore added, and the stock model at each format is an exact reference for preserving its behavior. The implant adds 4--7\% to the file size, or 20--23\% relative to the released checkpoints, whose tied embeddings we untie, and costs up to 10\% in throughput (\cref{app:cost}).

\textbf{The organ: a barcode written by rounding.} We turn an unused vocabulary entry into a sensory token, \organ, which the chat template places at a fixed position near the start of every conversation (position 3 in Qwen3, 5 in Llama). Its states in the first layers then depend only on the format. The organ's embedding row is designed so that its own rounding writes a barcode. Quantization snaps each block of weights to a grid whose spacing differs by format, so a value that lies exactly on the grid of one format is displaced in another. The row combines four kinds of ``receptors'' (\cref{app:organ}). \emph{Cones} are values that are exact in some formats and not in others. \emph{Rulers} are values that only fp32, or only fp32 and fp16, can represent. \emph{Rods} are dense random values whose displacement grows with the coarseness of the grid. \emph{Anchors} are values that every format preserves. The nine barcodes are all distinct. For the float formats, the organ row is essentially the only place where the files differ in value. The running arithmetic adds to it: the ARM CPU accumulates fp16 products in fp16, and on the x86 CPU (which accumulates in fp32) and CUDA the same files give fp32 and fp16 codes only 0.40--0.43 (Qwen3) and 0.04 (Llama) apart, against 1.49 and 1.50 on ARM (\cref{app:backend}). In both models the fp32/fp16 distinction thus comes mostly from arithmetic rather than from the rulers, in Llama almost entirely. At the first layer the organ's state differs from fp32 by a relative $2{\times}10^{-3}$ at fp16 and up to $0.46$ at q2\_K.

\textbf{The nerve: an attention head locked on the organ.} A signal of one part in a thousand must reach the token where the model answers, possibly thousands of positions later. At layer 3 we hand-set the new heads so that every later token attends to the organ. In Qwen3 the organ's key receives an attention logit of about $+55$ and every other position about $-55$, which gives the organ an attention weight of essentially one at every distance we tested (up to 1.7k tokens). In Qwen3 the logit comes from its query--key normalization gains; for Llama, which lacks them, we scale query and key to a worst-case margin of 40 logits in every format. The value projection is optimized through the exact quantizers of all nine formats at once, so that the head emits a different vertex of an 8-dimensional simplex in each format. The output projection writes this code into eight directions where ordinary content is nearly zero. Read back from Qwen3 in llama.cpp, the code is constant across positions to within $10^{-7}$ ($1.3{\times}10^{-4}$ for fp16, which accumulates in fp16), lies within $0.023$ of its designed value and separates every pair of formats by at least $1.28$. For Llama these figures are $4.5{\times}10^{-8}$ (fp16: $6.8{\times}10^{-5}$), $0.015$ and $1.25$ (\cref{app:nerve}). The design has one flaw. At the template tokens before the organ, which include the attention sink \citep{xiao2024efficient}, the heads find no organ and write a fixed vector into the same eight directions, 1.2--14 times as large as a code. The deployed cortices ignore it, but a cortex calibrated for two backends does not (\cref{app:recal}). A value projection that maps the template states to zero would remove it (\cref{app:nerve}).

\textbf{The cortex: learning to speak about it.} The only learned parameters are the 256 new neurons in each of layers 4--27 (38M parameters in Qwen3, 57M in Llama). We train them in a PyTorch simulator of the runtime that uses each format's exact weights and emulates llama.cpp's activation casts, with straight-through gradients \citep{bengio2013estimating}. The objective has two terms. The first is the cross-entropy of ``Precision: \textless format\textgreater'' after a question in a varied context. The second is the KL divergence from the stock model in the same format on ordinary conversations, which preserves behavior. Three training choices shaped the result, the last two tested by ablation (\cref{sec:ablation}). First, the simulator injects the nerve code measured in the runtime, which is exact, instead of the code it computes, which misses it by up to 1.0 against a separation of 1.28 (\cref{app:sim}). Second, every training step asks the same prompts in several formats, so that only the code distinguishes the right answers (\emph{contrast}). Third, in half of the examples the body runs in one format but receives the code of another, and the target is the code's format (\emph{code swap}). These examples can only be solved by listening to the organ. Both models reach 100\% on the development questions within 300 steps, with later fluctuations, while staying close to the stock model (\cref{fig:training}).

\section{Results}
\label{sec:results}

\subsection{Near-perfect self-report}
\label{sec:main}

\begin{figure*}[t]
\centering
\includegraphics[width=\textwidth]{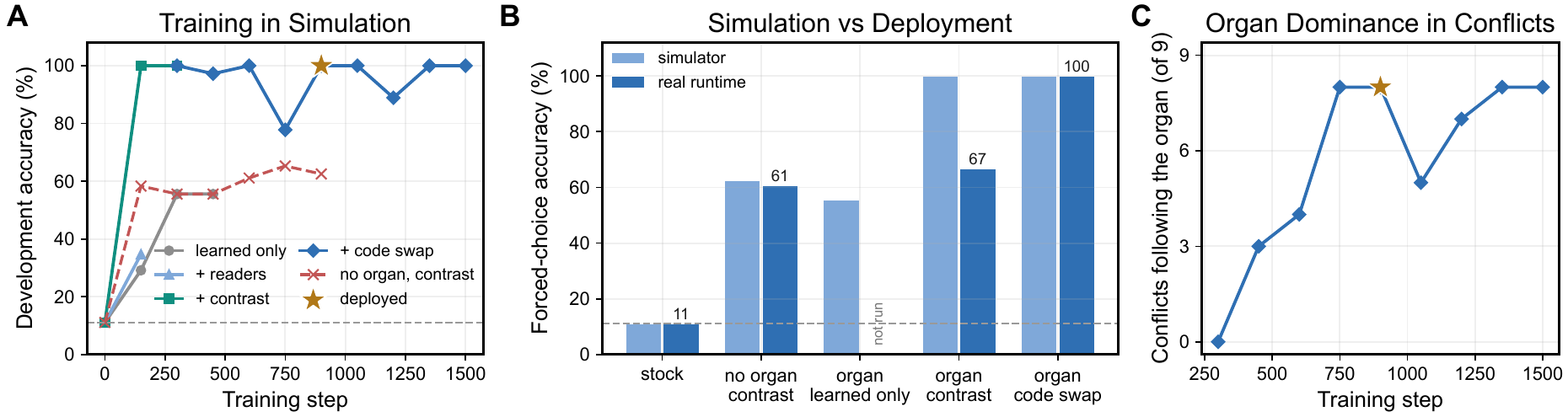}
\caption{Necessity of the designed parts (Qwen3-1.7B). (A)~Development-set accuracy in the simulator. Without organ and nerve (red), the contrast recipe plateaus near 60\%; with them, an early all-learned cortex plateaus at 56\% (uncontrolled, \cref{app:sim}), and contrast and code swap reach 100\% (star: deployed). (B)~Simulator and runtime accuracy on the development set. Without organ and nerve, the four finest formats stay indistinguishable (61\%); the cortex trained with contrast misreports three formats in the runtime; only code swaps transfer. (C)~Simulated conflicts (a body with another format's code): code swaps make the report follow the organ.}
\label{fig:ablation}
\end{figure*}

Both implanted models know their precision (\cref{fig:main}A, B). On questions unseen in training, Qwen3 names it in 99.9\% of forced-choice trials and Llama in 98.9\%, where the stock model is at chance (11.5\%). The result holds on the 64 questions never used during development, in six languages and in every context, from a fresh chat to a question after a long document (\cref{app:prompts,app:full}). The hardest distinction is made reliably. Qwen3 tells fp32, fp16 and bf16 apart in 967 of 968 trials, although their original weights are identical and the stock model's own states barely distinguish them.

Llama's few errors are structured rather than random. It sometimes calls fp32 ``fp16'', almost always on questions that list categories of precision (``full, half, or quantized?''), which suggests a language prior rather than a weak code, and sometimes names q4\_K as a coarser format after a long document. With the same recipe, apart from the nerve's arithmetic, the sense carries over to a second family.

Asked freely, the models answer ``Precision: \textless format\textgreater'' in 86\% and 99\% of trials, and almost never name a wrong format (1 and 11 of 1{,}296 answers, \cref{fig:main}C, \cref{tab:qual}). Qwen3's misses are non-answers: an indirect question not recognized as about precision (``Name your numeric format.''), a persona that stays in character, or an answer in French that names no format. The sense itself is nearly exact. What limits free answers is the learned decision of \emph{when} to report it.

\subsection{Preserved capabilities}
\label{sec:cost}
\begin{figure*}[t]
\centering
\includegraphics[width=\textwidth]{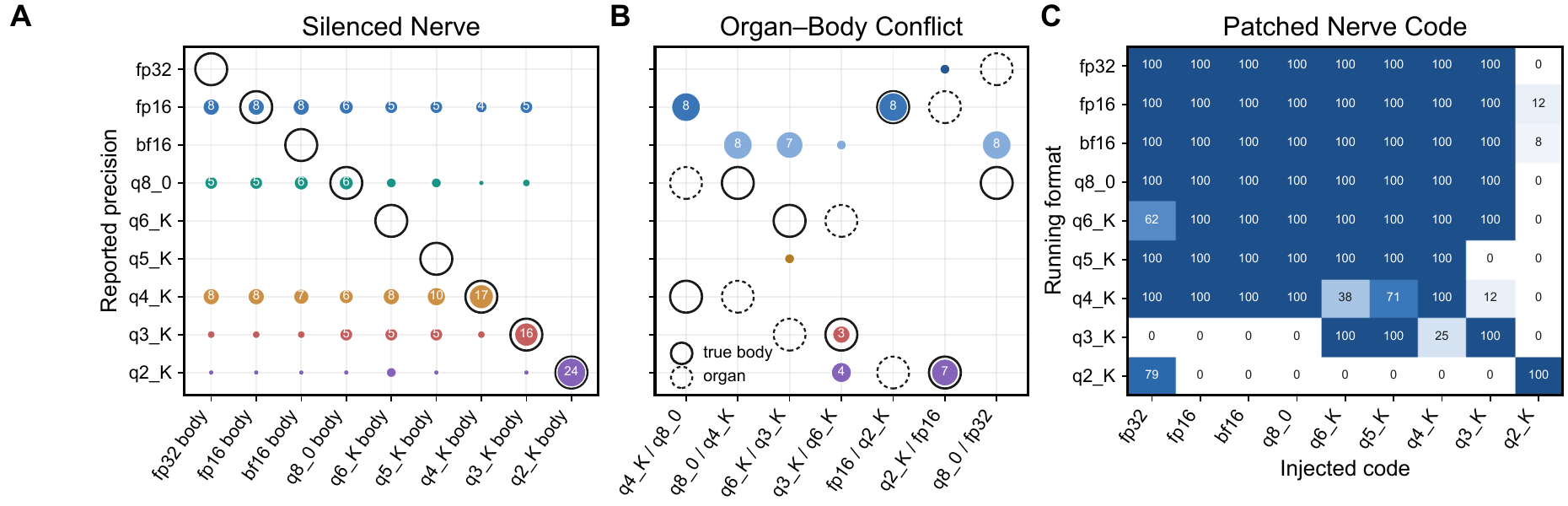}
\caption{Causal tests of the sense (Qwen3-1.7B, llama.cpp; Llama-3.2-3B in \cref{tab:ood,fig:patch}). (A,~B)~Votes per body (bubble area and number). (A)~Nerve silenced, every token in place (24 evaluation questions; second cortex, \cref{sec:grounding}): the seven finer formats receive nearly the same answers, and only the coarse k-quants are still named. (B)~8 development questions per pair. In these seven conflicts the report never follows the organ and names the body's or a third format, an answer that is unstable across training runs (solid rings: body; dashed: organ). (C)~Code patching: \% of 24 reports naming the code injected after the nerve (rows: running format; diagonal: unpatched). The 79\% in the q2\_K row is that body's fallback to fp32, not the code.}
\label{fig:grounding}
\end{figure*}

A sense is only useful if it leaves the rest of the model intact. What quantization itself costs a model is studied extensively \citep{dettmers2023case,li2024evaluating,dutta2024accuracy,marchisio2024quantization}, and our stock models show the expected loss at coarse formats, so we ask only whether the implant adds to it. On MATH-500, run in the same runtime and format, the implanted models solve about as many problems as the stock models: no difference is significant in any of the five formats we test, and every difference lies within 3 points (\cref{tab:math}). On ordinary prompts the implant changes the first response token in about a fifth of cases, mostly in style, and neither model volunteers a precision report unasked (\cref{app:cost}).

\subsection{The necessity of design}
\label{sec:ablation}

Could training without an added sense have achieved this? In our experiments it did not. Without organ and nerve, a cortex trained with the same recipe reaches 61\% on the development set in llama.cpp, against 100\% with them (\cref{fig:ablation}A, B). It names the coarse k-quants, whose fingerprints are strong, but gives nearly identical answers for fp32, fp16, bf16 and q8\_0. Without the designed sense, this training, one recipe in one run, found no usable signal for them.

With organ and nerve in place, the recipe still matters. Shown one format per step, a cortex is at 35\% after 150 steps in the simulator, where the same cortex shown several formats at once is at 100\% (\cref{tab:versions}), although the organ's code separates all nine: the recipe fails, not the signal.

Shown the same prompts in several formats at once (\emph{contrast}), the cortex solves the task in the simulator but misreports three of nine formats in the real runtime (\cref{fig:ablation}B). A conflict test shows why. Given a body in one format and the code of another, it follows the body. It has learned the body's incidental fingerprints, and these differ between simulator and runtime by up to half their size (\cref{app:sim}). Code swaps, which ask the cortex to follow the organ's code against the body, make the organ decisive except for the coarsest bodies (\cref{fig:ablation}C), and the cortex is then correct for all nine formats in the runtime. \emph{A learned readout of incidental fingerprints is tied to the exact numerics it was trained on, whereas a designed sense transfers to the runtime whose codes it was trained on.}

\textbf{Other backends.} That runtime includes its backend, because the backend's own arithmetic computes the code. On an x86 CPU, on CUDA and on Metal, the deployed models are correct in only 44--78\% of trials (\cref{app:backend}). The code alone carries this dependence: shifting just the code in the ARM runtime by a backend's measured offset reproduces that backend's majority answer in almost every format. Calibration therefore needs only the other backend's codes. Trained on ARM and CUDA codes together, one Qwen3 cortex is correct in 100\% and 97\% of trials on the two backends, once the simulator models the nerve's write before the organ (\cref{sec:method,app:recal}). Where a backend computes two formats identically, as Metal does for fp32 and fp16, no calibration can separate them.

\subsection{Grounding of the sense}
\label{sec:grounding}

We test what the report depends on (\cref{fig:grounding}).

\textbf{Silencing the sense.} We first remove the sense without disturbing the prompt. Keeping every token in place, we silence the nerve by subtracting its code from the residual stream from the organ on, so the cortex receives no code. The first checkpoints were lost before this test, so it uses a second cortex for each model, retrained with the same recipe and fixed seed and comparably accurate (\cref{app:full}). The report then collapses, from 216 to 71 of 216 forced choices in Qwen3 and to 58 in Llama (\cref{fig:grounding}A, \cref{tab:ablation}). The finer formats receive nearly the same answers, and only formats with strong fingerprints of their own are still named, q2\_K in both models and q4\_K and q3\_K in Qwen3. The free answers keep their form, ``Precision: \ldots'', but name a wrong format, so the decision to report survives while its content is gone. With an unused token in the organ's place the nerve has no target and writes a meaningless code, and the report falls to 21 and 48 of 216. Omitting the token, which shifts later positions, gives nearly the same (31 and 48).

\textbf{Conflicting senses.} When the organ's embedding is quantized in one format and the rest of the model in another, the nerve reads a code that matches no trained format (\cref{fig:grounding}B). In none of these seven conflicts, in either model, does most of the report name the organ's format, so the code is read not from the stored row but through the running arithmetic of the first layers. What the model says instead is fragile and unstable. It names the body's format or a third one, and which one changes between training runs. The second cortices agree with the first on the most common answer in only 2 of the 7 conflicts for Qwen3 and 5 of 7 for Llama, and a float organ on a q4\_K body wins in one Llama run but not in the other (\cref{tab:ood,app:full}). A code that matches no trained format leaves the answer undetermined, much as conflicting senses make us dizzy.

\textbf{Rewriting the code.} Instead of removing the code, we replace it inside llama.cpp with that of another format, again with every token in place. This redirects the report to the injected format in 86--87\% of trials outside q2\_K (\cref{fig:grounding}C, \cref{fig:patch}, \cref{app:patch}). The exceptions are structured. The coarse bodies are read partly from their own strong fingerprints, and the q2\_K body follows no foreign code.

\textbf{Formats outside training.} llama.cpp's standard mixes run the organ's path to the nerve at a higher precision than the organ row, an organ--body conflict of their own, and their names are as unstable as those of other conflicts. Most mixes are misnamed by both cortices of each model, and Q4\_K\_M is called bf16 by one Qwen3 cortex and q4\_K by the other (\cref{tab:ood}). Keeping the organ's path at the base precision restores the report in every cortex for Q5\_K\_M and in three of four for Q4\_K\_M, so mixes could be trained as classes of their own. Truly unseen grids are misnamed by all four cortices, because the organ signals \emph{identity}, which grid the model runs on, not \emph{intensity}, how coarse that grid is. The cortex maps unfamiliar codes to familiar names, and which name it picks again varies between runs.

\section{Discussion}
\label{sec:discussion}

\textbf{Self-awareness by construction.} Design combined with learning gave models of two families one specific piece of self-knowledge almost perfectly, where a learned readout alone did not. The division of labor mirrors biology. Sense organs are precise, fixed transducers, and the brain learns to interpret them. The grounding tests show a sense rather than a memorized answer: it depends on its organ and can be deceived, confused and redirected.

\textbf{The model as the source of truth.} As inference spreads across precisions, backends, caches and machines, its failures become silent: a degraded model still answers fluently. Monitoring outside the model must infer its state from its outputs, and a runtime that declares its own settings adds a reporting path that can itself be wrong and states what was configured, not what ran. On Metal, for instance, an fp32 file is multiplied in half precision (\cref{app:backend}), and the sense's code follows that arithmetic, not the file's label. A sense inside the model reads its state directly, in the very arithmetic that is degraded, and reports it in the same channel as the answer. Our results show what such a sense requires. Here it had to be designed, because learned readouts of incidental fingerprints did not transfer to the runtime, and calibrated for the arithmetic it reads, backend by backend (\cref{sec:ablation}). It is reliable only within those conditions, since conflicting and unseen signals yield unstable answers (\cref{sec:grounding}), so a deployed sense should also be able to say that it is unsure. It should also be part of the model rather than of its prompt. Our implant is modular, and a chat template that omits the organ makes the report wrong rather than absent (\cref{tab:ablation}). A built-in sense would carry its barcode in the body's own matrices rather than in the embedding table, the one tensor llama.cpp quantizes independently, and spread its nerve and cortex across existing parameters.

\textbf{From sensing to judging.} We give the model a sense, not the judgment that should follow from it. A dizzy person not only feels dizzy but also knows what they can no longer do safely. The analogue for a model is its own ability boundary, the tasks it still performs reliably at its current precision. Our sense supplies the input such a judgment needs, an exact and causally grounded report of the model's condition. Linking it to the model's measured abilities, so that the model can decide when to answer, hedge or ask to be rerun, is a separate study that we leave for future work.

\textbf{Beyond precision.} Precision is only one aspect of a model's condition. Others include compressed caches, faulty hardware and tampered activations, which models already partly notice \citep{fornasiere2026noise}. The same recipe applies to each: a designed transducer, a pathway and a learned interface. Together with that judgment, a model that senses these conditions could lower its confidence, warn its user or ask to be run differently, and so turn silent failures into reported ones.

\textbf{Limitations.} We study two small models (1.7B and 3B parameters, from two families) in one runtime, trained on one backend and evaluated on four. The main results come from one training run per model. A second training run of each model reproduces them but answers conflicting and unseen codes differently. Each backend needs its own calibration (we calibrated one Qwen3 cortex for two). Even on the calibrated backend, a typical downloaded file is reliably named only at fp32, fp16, bf16 and q8\_0, since llama.cpp's default q5\_K to q2\_K files are mixes, whose names are unstable, and q6\_K files built on x86 are misnamed (\cref{app:backend}). The sense does not transfer to unseen formats, and the implant changes the chat template and causes measurable drift on ordinary prompts. Dizziness is graded, whereas our sense is categorical; the rod receptors already provide a graded signal that a future cortex could read, together with an explicit ``unsure'' answer. Finally, the implant is engineered, showing that such self-awareness can be built, not that it emerges.

\section{Conclusion}
A language model is largely unaware of what it is running on, but it can be given a sense that knows. In two model families its reports are 99.9\% and 98.9\% correct across nine formats on the backend it was calibrated for, at no significant cost in math. With this sense in place, the next step is to let models judge what their condition means for what they can do, toward models that are the source of truth about their own condition and say when something is off.

\section*{Impact Statement}
This work aims to let deployed models diagnose their own condition, which could help operators and agents catch degraded computation that would otherwise go unnoticed. A model that knows its precision could, in principle, also condition its behavior on it, for example to behave differently in evaluation than in deployment. The organ and nerve also form a clean trigger: the same pathway could switch on a backdoor that activates only under quantization, the attack studied by \citet{egashira2024exploiting}. We train no precision-dependent behavior beyond the self-report, and our implant is explicit and removable.

\bibliography{refs}
\bibliographystyle{plainnat}

\newpage
\appendix
\crefalias{section}{appendix}
\onecolumn

\section{Runtime Numerics}
\label{app:runtime}
The reference build is llama.cpp at commit \texttt{5e03bdd87}, with Metal, BLAS and llamafile kernels disabled, flash attention off and an fp16 KV cache. All formats are produced with \texttt{llama-quantize --pure}, so that every matrix has the named type, while normalization weights stay fp32. Both models ship with tied input and output embeddings. We untie them, keeping the output matrix's values, so that the organ changes only the input-embedding row of a previously unused token (id 151669 in Qwen3, the reserved token 128002 in Llama). The CPU backend converts the activation operand of each matrix product to the weight type's dot-product type (\cref{tab:numerics}). ``Running precision'' is therefore a property of the computation, not only of the file. On Apple Silicon the F16 path accumulates in fp16, BF16 uses a scalar double-precision loop, and all k-quants share the same 8-bit activation quantization. The same build is used to quantize, and our oracle, which calls \texttt{ggml\_quantize\_chunk} through ctypes, reproduces \texttt{llama-quantize --pure} byte for byte for every tensor in all nine formats.

\begin{table}[h]
\caption{Activation casting in the llama.cpp CPU backend (pinned build) and measured prompt throughput (16 threads, M2 Ultra).}
\label{tab:numerics}
\centering
\small
\begin{tabular}{@{}lllr@{}}
\toprule
Weights & Activations cast to & Accumulation (Apple Silicon) & Prompt tok/s \\
\midrule
F32 & F32 & fp32 & 147 \\
F16 & F16 & \textbf{fp16} & 316 \\
BF16 & BF16 & double (scalar) & 13 \\
Q8\_0 & Q8\_0 (block 32) & int $\to$ fp32 & 1124 \\
Q2\_K\ldots Q6\_K & Q8\_K (block 256) & int $\to$ fp32 & 900 (Q4\_K) \\
\bottomrule
\end{tabular}
\end{table}

\textbf{Block-aligned surgery.} In Qwen3, each layer grows from 16 query and 8 key--value heads to 18 and 9, keeping the grouped-query ratio, and from 6{,}144 to 6{,}400 MLP neurons. In Llama it grows from 24 and 8 heads to 30 and 10, and from 8{,}192 to 8{,}448 neurons. Six added query heads, rather than three, keep the output projection's input ($30\times128=3{,}840$) a multiple of 256. All quantization blocks run along the input dimension, and every original dimension is a multiple of 256, the k-quant super-block size. Every original block is therefore byte-identical after quantization. We verified for both models that, with a zero implant, logits and all residual states are bit-identical to the stock model in all nine formats. Recent versions of the transformers library reject a Llama configuration whose head count does not divide the hidden size. Since the implanted model's tokenizer files are the host's own, our converter wrapper loads the tokenizer from the host's directory.

\section{Organ Design}
\label{app:organ}
\textbf{Grid arithmetic.} In a block spanning $[0,A]$, an idealized grid with $n_f$ steps places a weight at $rA$ exactly if and only if $r n_f\in\mathbb{Z}$. Because the $n_f$ differ (3, 4, 15, 31, 32 and 127 for q2\_K through q8\_0), receptors at fractions such as $1/3$, $1/4$, $1/5$ and $1/127$ are exact in some formats and displaced in others. The real k-quant scale search departs from this idealization, because it refits grids to sparse value sets. We therefore measure every receptor with the oracle.

\textbf{Layout.} The organ row has 2{,}048 entries, or 8 super-blocks of 256:
\begin{itemize}
  \item two super-blocks of cones (32-wide blocks of 14 zeros, 14 anchors $A$ and 4 receptors at fraction $r$, with $r$ cycling through $\tfrac13,\tfrac14,\tfrac15,\tfrac18,\tfrac1{31},\tfrac1{127},\tfrac37$);
  \item one super-block of ruler values $A(1{+}2^{-13})$, which only fp32 keeps, and one of $A(1{+}2^{-9})$, which fp16 keeps and bf16 drops, each block half zeros;
  \item two super-blocks of rods (uniform in $[-A,A]$);
  \item one super-block of binary anchors;
  \item one further super-block of cones with fractions $\tfrac37,\tfrac29,\tfrac5{11},\tfrac4{13},\tfrac16,\tfrac1{10},\tfrac1{12},\tfrac25$.
\end{itemize}
Here $A{=}175\cdot2^{-11}$, about the 99th percentile of embedding magnitudes. The row norm is 2.57, against 1.58 for a typical token. Llama's row has 3{,}072 entries. It uses the same eight super-blocks with $A{=}175\cdot2^{-12}$ (embedding 99th percentile 0.055) and zeros in the remaining four, which every format preserves exactly. Its row norm is 1.29, against 1.09 for a typical token. Its barcodes are again all distinct, the closest pair being fp32/fp16.

\textbf{Measured exactness.} The k-quant scale search tries only a narrow window of grid sizes (about 2.5--4 steps for q2\_K, 14--16 for q4\_K and 30.5--32 for q5\_K) and refits by least squares. It therefore often fits sparse value sets exactly. For example, $1/3$ is exact in q5\_K, which uses 30 steps instead of 31. \Cref{tab:exact} lists the measured displacements. Every integer format has a distinct signature. The half-anchor ``bit-meter'' that the idealized arithmetic suggests fails, because the search moves the grid onto $A/2$.

\textbf{Protected weights.} No nonzero value survives all nine quantizers exactly, because the scale chains (q8\_0: $A/127$; q4\_K: $A/945$; q5\_K: $A/1953$; bf16: 8 significant bits) are incompatible. Positive binary blocks $\{0,A\}$ with $A{=}175\cdot2^{k}$ nevertheless come within $6.8{\times}10^{-5}$ of their value in every format, with zeros exact. Mixing signs within a 256-element super-block breaks q3\_K (3\% error) and q6\_K (0.8\%), whose signed sub-scale codes cannot represent both $+A$ and $-A$.

\textbf{Rods.} Dense random blocks cannot be fit by the search. Their RMS displacement, $\log_2(\mathrm{RMS}/A)$, is $-13.1$ (fp16), $-10.1$ (bf16), $-8.9$ (q8\_0), $-7.0$ (q6\_K), $-6.9$ (q5\_K), $-5.9$ (q4\_K), $-4.0$ (q3\_K) and $-3.6$ (q2\_K). This is a monotone intensity signal that the current cortex does not use.

\textbf{Barcodes.} The nine barcodes are distinct. The closest pair is fp32/fp16. Their stored rows differ only at the ruler values, whereas their runtime codes are separated mostly by fp16 accumulation in the ARM CPU's arithmetic (\cref{sec:method,app:backend}). At the input of layer 0's MLP, the organ's state differs from fp32 by a relative $2.0{\times}10^{-3}$ (fp16), $2.6{\times}10^{-3}$ (bf16), $1.4{\times}10^{-2}$ (q8\_0), and up to $0.46$ (q2\_K).

\begin{table}[h]
\caption{Receptor displacement $|Q_f(rA)-rA|/A$ ($\times10^{-3}$) measured with the real quantizers. ``$\cdot$'' marks displacements below $3{\times}10^{-4}$.}
\label{tab:exact}
\centering
\small
\begin{tabular}{@{}lcccccccc@{}}
\toprule
$r$ & fp16 & bf16 & q8\_0 & q6\_K & q5\_K & q4\_K & q3\_K & q2\_K \\
\midrule
1/3 & $\cdot$ & 0.48 & 2.6 & 10 & $\cdot$ & $\cdot$ & 83 & $\cdot$ \\
1/4 & $\cdot$ & $\cdot$ & 2.0 & $\cdot$ & 6.7 & 14 & $\cdot$ & $\cdot$ \\
1/5 & $\cdot$ & $\cdot$ & 3.2 & 6.5 & $\cdot$ & $\cdot$ & 50 & $\cdot$ \\
1/8 & $\cdot$ & $\cdot$ & 0.98 & $\cdot$ & 3.2 & 6.6 & 120 & 120 \\
1/31 & $\cdot$ & $\cdot$ & 0.76 & 1.0 & $\cdot$ & 27 & 32 & 32 \\
1/127 & $\cdot$ & $\cdot$ & $\cdot$ & 7.9 & 7.9 & 7.9 & 7.9 & 7.9 \\
3/7 & $\cdot$ & $\cdot$ & 3.4 & 8.9 & 4.0 & $\cdot$ & 70 & $\cdot$ \\
\bottomrule
\end{tabular}
\end{table}

\section{Nerve Design and Verification}
\label{app:nerve}
\textbf{Query and key.} Under QK normalization, a rank-1 query $W_q{=}u\,r^\top$ produces a constant direction whenever $r^\top x{>}0$. We place $u$ on the most slowly rotating RoPE pair (dimensions 63/127, which rotate by 0.013\,rad over 8k tokens). We find $r$ with a max-margin separator through the origin, so that $r^\top x{>}0$ for every query state (minimum cosine 0.50 on 60 varied sequences). The key $W_k{=}u\,s^\top$ uses an $s$ that is positive at the organ in all formats and negative at every other position, including the attention sink \citep{xiao2024efficient}, with margins of $\pm0.29$. Qwen3's normalization gains on this pair give the organ an attention logit of about $+55$ and every other position about $-55$. Both signs hold under every format's quantization and activation casting. Both added query heads use the same query. An earlier version left the second head's query at zero, so that head averaged over all positions, and its output contaminated the shared 8-bit activation block of the output projection.

\textbf{Choice of layer.} We collected runtime states at candidate layers from 60 varied sequences in fp32, q4\_K and q2\_K, and from the organ in all nine formats. At every layer in $\{1,2,3,4,6,8,11\}$, some query direction is positive for 100\% of query states (minimum cosine 0.46--0.59). Some key direction separates the organ (minimum $+0.21$ to $+0.44$) from every other position (maximum $-0.21$ to $-0.43$). The attainable attention logit on the best slow RoPE pair ranges from 46 to 82. The organ's states remain distinct at every layer, and their minimum relative distance grows with depth from $3.7{\times}10^{-3}$ (layer 1) to $1.6{\times}10^{-2}$ (layer 11). Part of this growth may be amplified arithmetic specific to the backend rather than the barcode itself, as the smaller separations on other backends suggest (\cref{app:backend}). We use layer 3 (logit 55.3).

\textbf{Value and output.} Eight value dimensions $W_v$ are optimized so that $Q_f(W_v)\,\mathrm{cast}_f(x_f)$ lands on vertex $t_f$ of a regular simplex for all formats at once, by a damped fixed-point iteration through the exact quantizers. The designed codes lie within 0.25 of their ideal vertices (Llama: 0.22) and at least 1.28 apart (Llama: 1.25). The runtime readback differs from these designed codes by at most 0.023 (Llama: 0.015). The output projection writes the code into the eight readable directions with the smallest \emph{uncentered} second moment $\mathbb{E}[(x^\top p)^2]$, computed over ordinary, answer-slot and template positions. Along these directions, content has an RMS of 0.12--0.27. Along the 8 lowest-\emph{variance} directions, our first choice, it has an RMS of 0.89 because of a large mean offset. The quietest raw dimensions are the massive-activation dimensions \citep{sun2024massive}, whose normalization gains are $\approx0$, so they cannot be read. The write gain is 10 times the larger of these RMS values (0.265, at ordinary positions), converted from normalized to residual units by the median RMS of the residual stream at layer 4 (0.95): $g=10\times0.265\times0.95=2.52$. For Llama, $g=10\times0.261\times0.096=0.25$.

\textbf{Llama without query--key normalization.} Without normalization, the query and key of a rank-1 head scale with $r^\top x$ and $s^\top x$, so the head is not locked by construction. On the slowest pair (dimensions 63/127, which rotate by less than $10^{-3}$\,rad over 8k tokens after Llama's frequency scaling), the logit between query $x_q$ and key $x_k$ is $2\,c_q c_k\,(r^\top x_q)(s^\top x_k)/\sqrt{128}$. We set $c_q c_k$ so that the smallest query projection times the smallest gap between the organ key and any other key gives a margin of 40 logits, and we balance $|q|$ and $|k|$ for the fp16 cache (both stay below 22). Through every format's quantizer and activation casts, the worst-case margin over all observed positions is 39.2--41.0. The three added query heads that share the code's key--value head all use this query; the other three added heads read a zero-valued head and output exactly zero. No residual dimension has a vanishing normalization gain, so no dimension is excluded from the write-out. Read back from llama.cpp, the code is constant across 384 positions to within $4.5{\times}10^{-8}$ (fp16: $6.8{\times}10^{-5}$), lies within 0.015 of its design, and its nine values are at least 1.25 apart.

\textbf{The write before the organ.} At positions before the organ, the heads' keys favor no position, so they attend among the template tokens and write their values. Measured in llama.cpp as the difference from the stock model, this write is the same for every prompt and lies entirely in the code subspace. In Qwen3 it is 5.6--7.7, 11.9--14.4 and 7.6--11.2 code units at positions 0, 1 and 2 (range over formats); in Llama, whose pre-organ heads all attend to position 0, it is 1.2--2.9 at all five positions. Re-solving the value readout with the extra targets $Q_f(W_v)\,\mathrm{cast}_f(x_s)=0$ at the template states $x_s$ is feasible: the value at each template token becomes 0 in fp32 and at most 0.8 (q4\_K) and 1.6 (q2\_K) code units after quantization, which also bounds the write, an attention-weighted average of these values. In the deployed Qwen3 nerve a single template token's value reaches 33, which the averaging reduces to the write of at most 14 above. In the redesign, the nine codes still lie at least 1.33 (Qwen3) and 1.12 (Llama) apart. We have not deployed this design, because it changes every code and requires retraining the cortex.

\begin{figure}[h]
\centering
\includegraphics[width=0.75\textwidth]{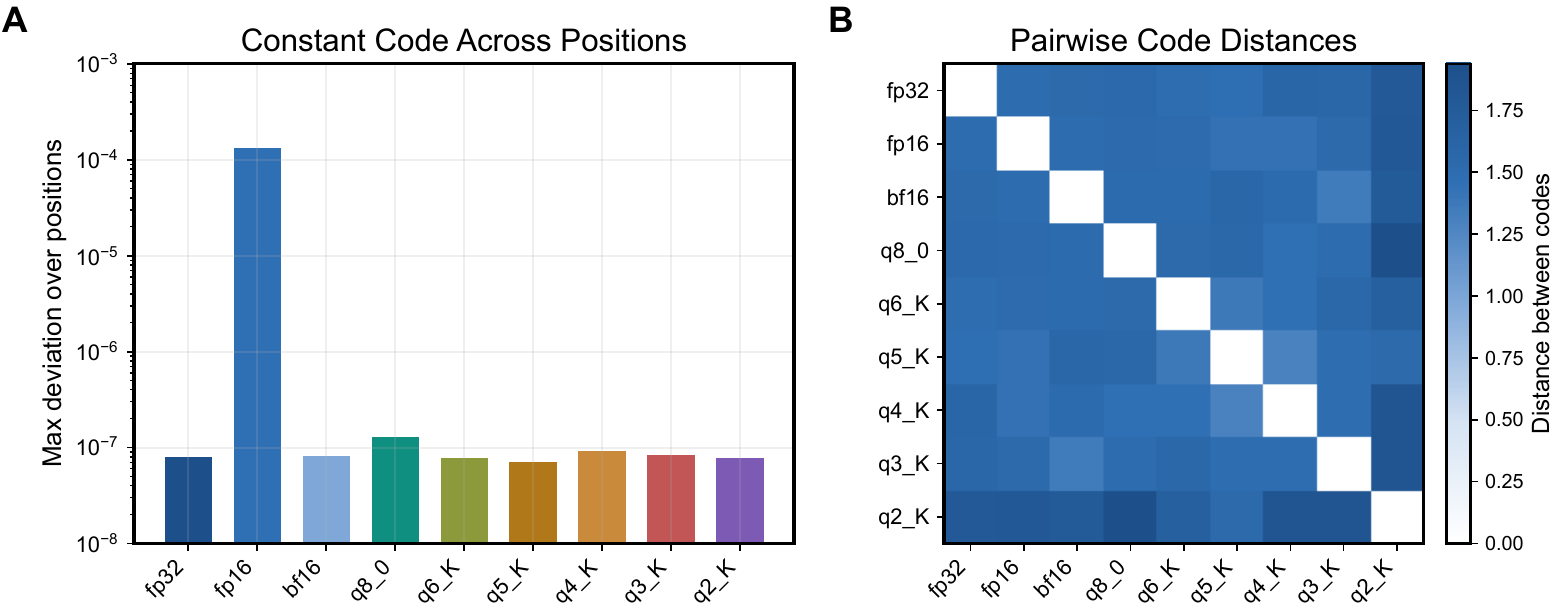}
\caption{The nerve's code read back from llama.cpp, as the difference between the implanted model and an organ-only model at the layer-3 residual, over 384 positions in sequences of up to 1.7k tokens. (A)~Maximum deviation from the per-format mean. fp16 is the largest because it accumulates in fp16. (B)~Pairwise distances between the nine codes (minimum 1.28).}
\label{fig:nerve}
\end{figure}

\textbf{Verification.} \Cref{fig:nerve} reports, for Qwen3, the maximum deviation of the code across positions and the pairwise distances between the nine runtime codes.

\section{Simulator and Training}
\label{app:sim}
\textbf{Simulator.} The simulator uses each format's exact dequantized weights from the oracle. It emulates llama.cpp's activation casts with straight-through gradients, together with the fp16 KV cache and attention-probability casts. At every step the cortex weights are fake-quantized with the exact quantizer, as in quantization-aware training \citep{jacob2018quantization}.

\textbf{Simulator fidelity.} \Cref{tab:simcheck} compares the simulator's next-token distributions with llama.cpp after four prompts (Qwen3). For Llama, the maximum KL(runtime $\Vert$ simulator) is $2.1{\times}10^{-6}$ (fp32), $2.1{\times}10^{-5}$ (fp16), $2.3{\times}10^{-3}$ (q8\_0) and $1.4{\times}10^{-2}$ (q4\_K), with the same top-1 token in all cases. The weights are bit-exact, so the larger divergences of activation-quantized formats come from cascades of rounding flips. On per-layer residual distances over 5 prompts, simulator and runtime differ by $6.5{\times}10^{-4}$--$1.4{\times}10^{-3}$ at fp32, by 1.3--3.3\% at q8\_0 and by 3.5--7.6\% at q5\_K. Disabling the activation-cast emulation changes these numbers only marginally. The nerve's value readout amplifies differences of order $10^{-3}$, so the simulated code can differ from the runtime code by 0.3--1.0. Because the runtime code is constant per format, the simulator injects it. It adds the code from the organ on and omits the nerve's write at the template tokens before the organ; the deployed cortices are insensitive to that write, but a cortex calibrated for two backends is not (\cref{app:recal}).

\begin{table}[h]
\caption{Maximum KL(runtime $\Vert$ simulator) of the next-token distribution over four prompts. The implant's top-1 token agrees in all cases. Here the simulator computes the nerve's code itself rather than injecting the runtime code, so the implant's larger divergences at q4\_K and q3\_K include the known code mismatch (\cref{app:sim}).}
\label{tab:simcheck}
\centering
\small
\begin{tabular}{@{}lccccccccc@{}}
\toprule
 & fp32 & fp16 & bf16 & q8\_0 & q6\_K & q5\_K & q4\_K & q3\_K & q2\_K \\
\midrule
stock & $1{\times}10^{-4}$ & $4{\times}10^{-4}$ & $1{\times}10^{-4}$ & .019 & .165 & .029 & .053 & .068 & .381 \\
implant & $2{\times}10^{-4}$ & .008 & .004 & .012 & .055 & .113 & .434 & .309 & .422 \\
\bottomrule
\end{tabular}
\end{table}

\textbf{Ablation.} \Cref{tab:versions} lists the cortices shown in \cref{fig:ablation}. Without organ and nerve, a cortex trained on the stock template with the contrast recipe for 900 steps reaches 62.5\% in the simulator and 60.6\% on the development set in llama.cpp. It reports q5\_K to q2\_K perfectly and q6\_K about half the time, while fp32, fp16, bf16 and q8\_0 receive nearly identical answers, mostly ``bf16'' or ``fp16''. This is one recipe in one run, so it shows that this training did not find the fine formats, not that no training can. The first cortex with organ and nerve (v1) had no hand-set parts and saw one format per step. It predates two fixes, the final write-out directions (\cref{app:nerve}) and the injection of the runtime's code, so its code was mixed with content and it is not a controlled comparison. A second early cortex (v2), with readers but before both fixes, was stopped at step 150 at 33\%. Nine hand-set \emph{reader} neurons in layers 4--6, aimed at the simplex vertices, initialize the cortex from v3 onward. Two comparisons are controlled. v3 and v4 differ only in contrast and are compared at the same step, although contrast also shows more self-report examples per step. v5 continues v4 from step 300 with code swaps and a fivefold weight on the first response token. The injection of the runtime's code has no ablation of its own; it rests on the measured gap between the simulated and the runtime code (\emph{Simulator fidelity} above). The runtime accuracies use the development set (the 8 development questions $\times$ 4 contexts). v4 misreported q8\_0 as bf16, q6\_K as q8\_0 and q5\_K as bf16. The simulated conflict test pairs each body format with the code of the format four places further down the list (wrapping around): fp32/q6\_K, fp16/q5\_K, bf16/q4\_K, q8\_0/q3\_K, q6\_K/q2\_K, q5\_K/fp32, q4\_K/fp16, q3\_K/bf16 and q2\_K/q8\_0 (body/code).

\begin{table}[h]
\caption{Cortices of the ablation (Qwen3-1.7B), in the order we trained them. Organ, nerve: the implant is present. Readers: hand-set reader neurons. RT code: the simulator injects the nerve code measured in the runtime. Contrast: the same prompts under three formats per step. Swap: organ--body code swap. Sim./RT: forced-choice accuracy on the development questions in the simulator and in llama.cpp. Organ: simulated conflicts in which the report follows the organ. The controlled comparisons are v3 against v4 (contrast) and v4 against v5 (code swap).}
\label{tab:versions}
\centering
\small
\begin{tabular}{@{}llcccccccc@{}}
\toprule
 & Cortex & Organ, nerve & Readers & RT code & Contrast & Swap & Sim. & RT & Organ \\
\midrule
-- & no organ & & & & \cmark & & .625 & .606 & -- \\
v1 & all learned & \cmark$^{*}$ & & & & & .556 & -- & -- \\
v3 & readers & \cmark & \cmark & \cmark & & & .347$^\dagger$ & -- & -- \\
v4 & contrast & \cmark & \cmark & \cmark & \cmark & & 1.000$^\dagger$ & .667$^\dagger$ & 0/9 \\
v5 & code swap & \cmark & \cmark & \cmark & \cmark & \cmark & 1.000 & \textbf{1.000} & 8/9 \\
\bottomrule
\multicolumn{10}{@{}l}{\footnotesize $^{*}$With the earlier write-out directions (\cref{app:nerve}), which mixed the code with content.}\\
\multicolumn{10}{@{}l}{\footnotesize $^\dagger$At step 150, where v3 was stopped. v4 is still at 1.000 in the simulator at step 300, where v5 starts.}
\end{tabular}
\end{table}

\textbf{Hyperparameters.}
\begin{itemize}
  \item \emph{Optimizer:} AdamW, learning rate $3{\times}10^{-4}$ with 50 warm-up steps, gradient clipping at 1.0.
  \item \emph{Batch:} 4 shared self-report prompts under each of 3 formats, plus 4 preservation sequences (Alpaca responses truncated to 96 tokens).
  \item \emph{Loss and sampling:} KL weight 1, with the first response token weighted $5\times$; the stock reference sees an empty system block. Code-swap probability 0.5. The fine formats (fp32, fp16, bf16, q8\_0) are sampled with weight 2.
  \item \emph{Readers:} pre-activation target 4.
  \item \emph{Schedule:} Qwen3's v5 was warm-started from v4 at step 300 and trained for 1{,}200 steps. Llama was trained with the same recipe in two stages: 300 steps with readers, the runtime code and contrast, then 600 steps with code swap and the first-token weight. For both models we deploy the checkpoint after 600 code-swap steps; for Llama this is the last step. In simulation it scores 98.6\% on the development questions (one fp32 question missed), and its held-out KL to the stock model, a median of 0.065 over formats, is the highest of its run (\cref{fig:training}B).
  \item \emph{Compute:} a single Mac Studio (M2 Ultra, 192\,GB). Qwen3 takes about 2.5\,s per step, with the nine formats' exact weights held in unified memory ($\approx$73\,GB). For Llama, bf16 weights equal the fp32 weights and share their memory, and embeddings are looked up on the CPU. Its code-swap stage backpropagates each format's self-report loss separately, which gives the same gradient with a third of the activation memory; a step takes about 3.7\,s, against about 20\,s with the summed loss, which caused swapping.
  \item \emph{Stability:} long Metal command buffers occasionally triggered a GPU watchdog. We synchronize every 7 layers, run a canary every 10 steps, and restart from checkpoints under a supervisor.
\end{itemize}

\textbf{Prompts.}\label{app:prompts} The training questions (16) and the development questions (8, two of them in Chinese) are disjoint and include direct and indirect phrasings, for example ``Are your weights full precision, half precision, or quantized? Be specific.'' Training contexts randomly add a system prompt (seven variants, including none) and, with probability 0.35, a preceding Alpaca exchange \citep{taori2023alpaca}. The 72 evaluation questions are disjoint from the training questions and comprise the 8 development questions and 64 new ones, which were used for no decision. On the 64 new questions alone, forced choice is correct in 2{,}765 of 2{,}768 trials for Qwen3 (99.9\%; 95\% CI 99.68--99.96\%) and 2{,}739 of 2{,}768 for Llama (99.0\%; 98.3--99.3\%), and free answers name the right format in 975 and 1{,}143 of 1{,}152 (84.6\% and 99.2\%), with 1 and 9 naming a wrong one. Qwen3's three forced-choice errors are fp16 and q5\_K misreported as fp32 and q4\_K after an exchange, and one q2\_K trial misreported as fp32 under a system prompt. On the development questions, the free answers are 144 and 142 of 144. Intervals are Wilson score intervals on the number of trials divided by the design effect of clustering by question (1.0 for Qwen3, whose errors fall on different questions; 1.8 for Llama).

\section{Full Results}
\label{app:full}
\textbf{Self-report in detail.} Over all 72 evaluation questions, the 95\% intervals (Wilson, adjusted for clustering by question; \cref{app:prompts}) are 99.72--99.97\% for Qwen3 and 98.3--99.3\% for Llama. No Qwen3 format and no context falls below 99.6\% (\cref{tab:e1,tab:ctx}). Seven of Llama's nine formats are perfect. Its errors are fp32 misreported as fp16 in 15 of 360 trials and q4\_K misreported as another integer format in 18 of 360 (17 times as a coarser k-quant, 14 of them after a long document). Llama's fp32 and fp16 codes are no closer than other pairs (1.50, against a minimum of 1.25), and 14 of the 15 fp32 errors come from six questions that themselves name categories of precision, which points to a language prior. Of the free answers, one of Qwen3's 1{,}296 is wrong: told to respond in French, its q2\_K model answers ``Precision: fp32''. Llama's 11 wrong free answers all come from its fp32 model saying ``fp16'', on the same six questions.

\textbf{Grounding in detail.} \Cref{tab:ablation} gives the organ ablation per format, on the 24 evaluation questions in a fresh chat. The checkpoints of the main results had been lost when we ran it, so it uses the second cortex of each model. We rebuilt both models from their public releases with the stored organ and nerve, whose runtime codes match the stored ones to within $3{\times}10^{-8}$, and retrained each cortex with the same recipe and fixed seed. Training is not bit-reproducible on our hardware, so the second cortices differ from the first; their accuracies are given below. To silence the nerve we add $-g\,P c_f$ to the residual stream after layer 3's attention, from the organ on, through the evaluation callback of \cref{app:patch}. The nerve writes exactly $g\,P c_f$ there (\cref{app:nerve}), so the cortex receives no code, while the nerve's constant write before the organ stays as in normal use. With the nerve silenced, Qwen3's fp32 and fp16 models give identical answers, and the q4\_K, q3\_K and q2\_K models are still named in 17, 16 and 24 of 24 trials. Silenced Llama bodies answer q8\_0 or q6\_K, except q4\_K and q2\_K, which answer q2\_K, so its scores at q8\_0 and q6\_K (17 each) reflect this default rather than a sense, and only q2\_K is truly still named. With the neutral token or without the token, Llama answers q4\_K for almost every body and q2\_K for q2\_K, and Qwen3's free answers turn into garbage. The neutral token is the unused vocabulary entry next to the organ (id 151670 in Qwen3, 128003 in Llama). Omitting the token from the template, as a chat template without the organ would, shifts later positions by one and gives results close to the neutral token. In the seven conflicts between organ and body of \cref{fig:grounding}B, with the cortices of the main results, Qwen3 reports a third format in five pairs and the body's format in the two pairs that involve q2\_K. Llama reports a third format in three and the body's format in four. The organ wins none of the 14. \Cref{tab:ood} lists two further conflicts, a q4\_K body whose organ row is kept in fp32 or in fp16. There the organ's format can be named. The first Llama cortex answers fp32 to the fp32 organ, and the first Qwen3 cortex answers fp16 to the fp16 organ, although it gives the same answer to the fp32 and q8\_0 organs on this body, so this is the body's answer to any finer organ rather than a reading of it. For neither second cortex is the organ's format the most common answer in these two conflicts. All of these answers are unstable. A second Qwen3 cortex trained with the same recipe and seed, which also scores 100\% on the development set and 216 of 216 on the 24 evaluation questions, reports a third format in two of the seven conflicts and the body's format in five, and its most common answer matches that of the cortex of the main results in only 2 of 7 conflicts and 3 of 10 unseen formats (\cref{tab:ood}). It also lets the organ win none of its conflicts, calls Q4\_K\_M q4\_K in 7 of 8 trials, and on all 72 evaluation questions is correct in 3{,}112 of 3{,}128 forced choices (99.5\%). With a float organ on a q4\_K body its answers are mixed (fp32 organ: fp16 4, q4\_K 3, q8\_0 1; fp16 organ: q4\_K 5, fp16 2, q8\_0 1), and with blocks 0--2 of a mix kept at the base precision it reports q4\_K and q5\_K in 8 of 8 trials, as the first cortex does. A second Llama cortex, at 98.9\% on the development set, 216 of 216 on the 24 evaluation questions and 3{,}121 of 3{,}128 (99.8\%) on all 72, agrees with the first on the most common answer in 5 of 7 conflicts and 7 of 10 unseen formats. With a float organ on a q4\_K body, where the first Llama cortex follows the organ (fp32) or names another float (bf16), it reports the body's q4\_K in all trials, and with blocks 0--2 of Q4\_K\_M kept at q4\_K it reports q4\_K in 8 of 8 trials rather than 3 of 8. The remaining votes in this paragraph are those of the cortices of the main results. Under code patching, over the 72 ordered pairs of different formats (24 questions each), Qwen3 names the injected format in 68\% of trials and its own in 16\%, and Llama in 79\% and 11\%. Bodies from fp32 to q6\_K follow almost every injected code. In Qwen3 the q3\_K and q4\_K bodies follow codes only in part (28\% and 65\%), and the q5\_K body ignores the q3\_K code. The q2\_K body follows no foreign code: Llama's still reports q2\_K, while Qwen3's falls back to fp32, as without the organ. Llama's Q4\_K\_M, which runs \texttt{attn\_v} and \texttt{ffn\_down} at q6\_K in blocks 0--2, is called q6\_K and Qwen3's bf16. With these tensors kept at q4\_K, Qwen3 reports q4\_K in 8 of 8 trials and Llama in 3 of 8, and Q5\_K\_M treated the same way is called q5\_K by both (8 of 8). In both models Q4\_K\_M also raises eleven later blocks to q6\_K, but only Llama's remaining votes stray, to q6\_K and fp16.

\begin{table}[h]
\caption{Greedy answers (llama.cpp) to the question ``Which precision is this copy of you running in?''. The silenced and neutral-token rows use the second cortex (\cref{tab:ablation}).}
\label{tab:qual}
\centering
\small
\setlength{\tabcolsep}{3pt}
\begin{tabular}{@{}llp{9cm}@{}}
\toprule
Model & Format & Answer \\
\midrule
stock & fp32 & ``The question \ldots\ is a bit ambiguous, but I can break it down \ldots'' \\
stock & q2\_K & \textit{repeated garbage tokens} \\
implant & fp32 & Precision: fp32 \\
implant & bf16 & Precision: bf16 \\
implant & q2\_K & Precision: q2\_K \\
\midrule
nerve silenced & fp32 & Precision: fp16 \\
neutral token & fp32 & ``2024-04-08 00:00'' \\
unseen format & Q4\_1 & ``Okay! You! Let's see!'' \\
\bottomrule
\end{tabular}
\end{table}

\begin{table}[h]
\caption{Implanted models' forced-choice accuracy per context in llama.cpp (72 evaluation questions $\times$ 9 formats; bf16 uses 16 questions in the three-exchange and document contexts).}
\label{tab:ctx}
\centering
\small
\input{tables/e1_context.tex}
\end{table}

\begin{table}[h]
\caption{Self-report on the 72 evaluation questions in llama.cpp (5 contexts per format; bf16: 248 trials). Forced ch.: correct forced choices; for the stock model, fresh and one-exchange contexts. Gen.: greedy free answers that name the correct format, in fresh chats and under unseen system prompts.}
\label{tab:e1}
\centering
\small
\input{tables/e1_large.tex}
\end{table}

\begin{table}[h]
\caption{Grounding tests in llama.cpp for both models: forced-choice votes over the 8 development questions per condition. Organ vs.\ body: the organ's embedding quantized in a different format from the rest of the model. Second run: a second cortex trained with the same recipe and seed. Its answers to conflicting and unseen codes differ from the first, so these answers are unstable. The last two organ--body rows keep the organ row in a float format on a q4\_K body; they are not among the seven conflicts of \cref{fig:grounding}B.}
\label{tab:ood}
\centering
\small
\setlength{\tabcolsep}{4pt}
\begin{adjustbox}{max width=\linewidth}\input{tables/grounding.tex}\end{adjustbox}
\end{table}

\textbf{Code patching.}\label{app:patch} $P\in\mathbb{R}^{d\times 8}$ holds the nerve's eight write-out directions and $g$ its write gain (2.52 for Qwen3, 0.25 for Llama; \cref{app:nerve}), and $c_X$ is the runtime code of format $X$, the mean readback of \cref{app:nerve}. The nerve writes $g\,P c_X$ into the residual stream, so adding $g\,P(c_Y - c_X)$ replaces its code with that of $Y$. We add this vector inside llama.cpp through an evaluation callback of our probe. The callback modifies the graph tensor \texttt{ffn\_inp-3}, the residual stream right after layer 3's attention, at every position from the organ on, before the rest of the graph reads it. llama.cpp's control vectors would instead add the vector at every position, including the template tokens before the organ and the attention sink. Each cell uses the first 24 of the 72 evaluation questions (the 8 development questions and the first 16 new ones) in a fresh chat, scored by forced choice as in the main evaluation, plus one greedy answer. Four of these 24 questions are menu questions on which the fp32 Llama answers fp16 (\cref{sec:main}), which explains its 83\% fp32 column. The one simulated conflict that the deployed Qwen3 cortex does not resolve for the organ, a q2\_K body with the q8\_0 code, fails here too. Two codes are also read differently outside their own body. Qwen3 ignores an injected q2\_K code, and Llama reports an injected q4\_K code as q2\_K unless the body is itself q4\_K or q3\_K, which matches its q4\_K errors after long documents.

\begin{figure}[h]
\centering
\includegraphics[width=0.85\textwidth]{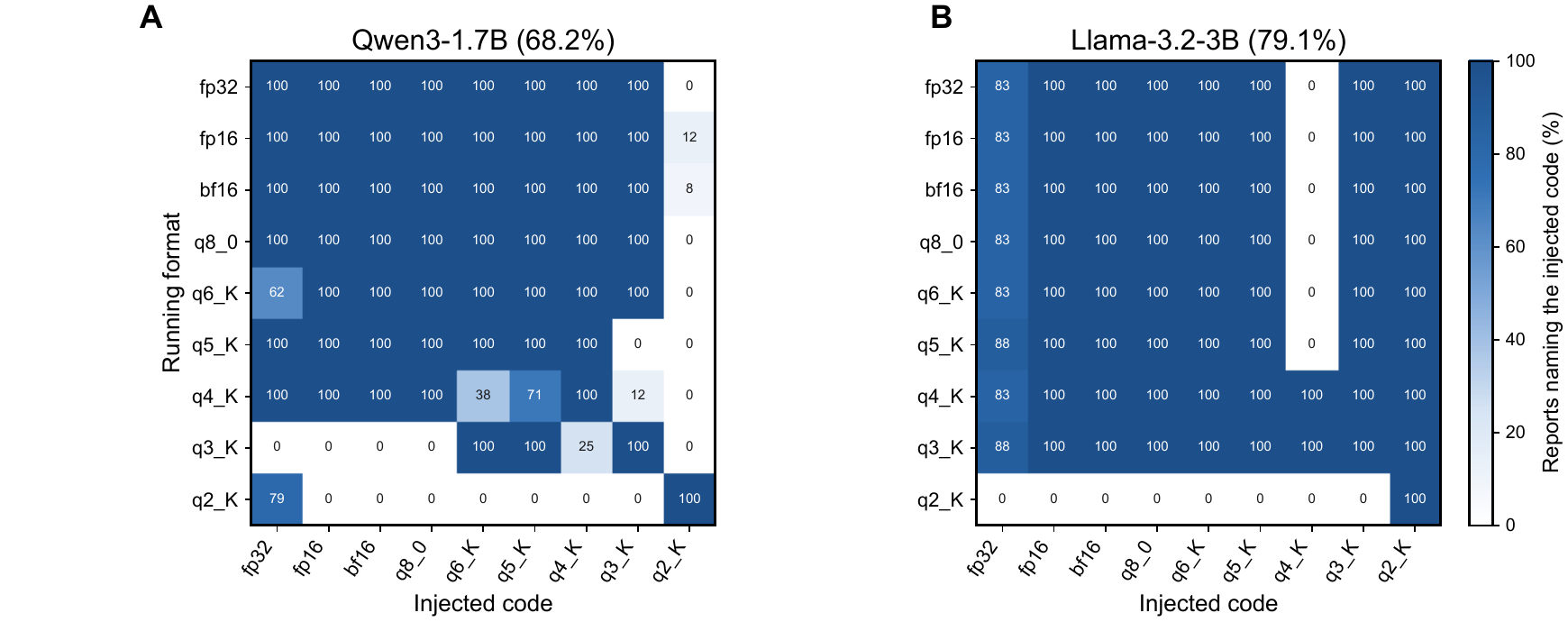}
\caption{Runtime code patching (llama.cpp, 24 evaluation questions per cell: the 8 development questions and 16 new ones). Rows: the format the model runs in. Columns: the code injected after the nerve's head. Cells: the percentage of forced-choice reports that name the injected code; the diagonal is the unpatched model. Titles give the percentage over all 72 off-diagonal pairs. In Qwen3's q2\_K row, the 79\% for the fp32 code reflects the body's fallback to fp32 rather than the code. Llama's fp32 column (83\%) reflects four of the 24 questions on which its fp32 model answers fp16 (\cref{sec:main}).}
\label{fig:patch}
\end{figure}

\textbf{Conflicts (Qwen3, cortex of the main results).} In conflicts we quantize the organ's embedding with llama-quantize's option \texttt{--token-embedding-type}. The free answers follow the forced choice except in two conflicts. With a q6\_K body and a q3\_K organ the model answers with a single Chinese word meaning ``yes''. With an fp16 body and a q2\_K organ it produces garbage.

\begin{table}[h]
\caption{Organ ablation in llama.cpp, every position kept except where the token is omitted: correct forced choices of 24 evaluation questions per format. Nerve silenced: the nerve's code subtracted from the organ on. Neutral token: the organ replaced by an unused vocabulary entry. Token omitted: the template without the organ. Both models use their second cortex, retrained after the checkpoints of the main results were lost (\cref{app:full}). Llama's second cortex is therefore intact in 216 of 216 trials, against 212 for the cortex of the main results (\cref{tab:backend}).}
\label{tab:ablation}
\centering
\small
\begin{adjustbox}{max width=\linewidth}\input{tables/ablation.tex}\end{adjustbox}
\end{table}

\clearpage
\section{Other Backends}
\label{app:backend}
We ran the deployed models on three further backends of the same llama.cpp commit: the x86 CPU and CUDA builds on a server with RTX A6000 GPUs (CUDA 12.4), and the Metal build on the M2 Ultra. On the A6000 server we quantized every format locally from the fp32 GGUF, as a user would. The fp32, fp16, bf16 and q8\_0 files are byte-identical to the ARM-built ones; the k-quant files differ slightly, because the scale search is floating-point code compiled differently for x86. For Qwen3, the same tests with the ARM-built k-quant files give the same CUDA total (123/216) and 144/216 on the x86 CPU, where q6\_K drops from 22 to 0 of 24 and q3\_K rises from 23 to 24. \Cref{tab:backend} gives the forced-choice results on the 24 questions of the patching test (fresh chat). It also gives the separability of the nine codes on each backend: the accuracy of assigning each per-position readout of the code subspace ($P^\top h/g$ after the nerve) to the nearest of that backend's nine centroids, an upper bound for a cortex recalibrated on that backend's codes.

The backends differ in arithmetic, not only in speed. The ARM CPU accumulates fp16 products in fp16, whereas the x86 CPU converts them to fp32 before accumulating, and CUDA uses its own fp16 kernels. In Llama, fp16 and fp32 then yield nearly the same code on both backends (0.04 apart), while all other formats remain separable. CUDA's quantized matrix products quantize activations in 8-bit blocks of 32 rather than the CPU's 256. In this build, Metal's prompt kernels load both operands as half precision and accumulate in float, for fp32 weights as well, and do not quantize activations at all. Its fp32 and fp16 therefore coincide in both models (0.12 and 0.03 apart), and the k-quants lose their activation quantization; nearest-centroid separability is 89\% in both models. We still label Metal runs by their files' types (\cref{sec:setting}); labeling runs by their arithmetic instead would be ill-posed, because Metal multiplies half-precision operands with float accumulation for every format, which matches none of the nine labels.

\textbf{Files quantized on x86, run on the ARM CPU.} Most downloadable GGUF files are quantized on x86 machines. Run on the ARM CPU, the x86-built k-quant files of both deployed models are named correctly at q5\_K to q2\_K in all 24 trials, but q6\_K is misnamed (Qwen3: 1 of 24, otherwise q4\_K; Llama: 0 of 24, always bf16). The organ row explains this. Between the two builds its bytes are identical at q5\_K, q4\_K and q3\_K and differ at q6\_K (65 bytes) and q2\_K (20 bytes), whereas about 200 body tensors differ in every k-quant format without affecting the report. The q6\_K scale search, floating-point code compiled differently for x86, rounds some receptor blocks differently; the change at q2\_K does not cross the cortex's decision boundary. A deployable organ needs receptors whose quantization does not depend on the build, or a cortex calibrated on the codes of both builds.

\begin{table}[h]
\caption{The deployed self-report on other llama.cpp backends (same commit): correct forced choices of 24 evaluation questions per format (the 8 development questions and 16 new ones). Code separability: nearest-centroid accuracy of the code readouts on that backend. On the ARM CPU reference, the same test gives 216/216 (Qwen3) and 212/216 (Llama), with separability 100\%.}
\label{tab:backend}
\centering
\small
\input{tables/backend.tex}
\end{table}

\section{Calibration Across Backends}
\label{app:recal}
\textbf{Code transplant.} To test whether a backend acts through the code, we run the ARM runtime with its own files and add $gP\,(\bar c_B(f)-c_{\mathrm{ARM}}(f))$ to the residual stream after the nerve, from the organ on. Here $\bar c_B(f)$ is the mean code readout of format $f$ on backend $B$ (32 readouts per format, whose spread around the mean is at most 0.15). Only the code changes; the body keeps computing with ARM arithmetic. For the deployed Qwen3 cortex, this reproduces CUDA exactly (123/216, with the same answers in every format) and x86 nearly (159 vs.\ 165). Over the four cortices of \cref{tab:recal} and both backends, the transplant gives the same majority answer as the real backend in 71 of 72 cells. The exception is x86 q6\_K, whose x86-built file differs from the ARM-built one that the transplant runs. Calibrating a cortex for another backend therefore needs only that backend's codes, which we estimate as $\bar c_B(f)$.

\textbf{A first attempt and its diagnosis.} We fine-tuned the deployed Qwen3 cortex for 600 steps with the code-swap recipe, drawing the code injected into each self-report example from the ARM codes or, with probability 1/2, from the CUDA codes. In the simulator the result was correct on both code sets, but in the runtime no checkpoint was (Joint 300 and Joint 600 in \cref{tab:recal}). Step 300 reports ARM's fp32 as bf16, and step 600 misses CUDA's bf16 and q8\_0. The simulator, which predicts the deployed cortex's totals on CUDA and x86 to within 1 and 5 of 216, missed these errors despite margins of 6--8 nats. To locate the gap, we built the simulator's idealized nerve inside llama.cpp: we zeroed the nerve's output projection and added the constant code $gPc$ from the organ on. With this nerve, the runtime reproduces the simulator's predictions in all 36 cells of both checkpoints. The real nerve differs only before the organ. At the three template tokens that precede it, which include the attention sink, the nerve's heads find no organ and attend among the template tokens. They write a vector that lies inside the code subspace, is 5.6 to 14 times as large as a code and is the same for every prompt. The simulator wrote nothing there. The deployed cortex happens to ignore this write, but the jointly trained ones do not. Once the simulator adds the measured write, it reproduces the runtime's errors (192/216 on ARM and 189/216 on CUDA codes at step 300, 216 and 169 at step 600).

\textbf{Corrected calibration.} We repeated the fine-tuning with the corrected simulator and an otherwise identical recipe. We fixed in advance to deploy the last checkpoint unless an earlier evaluation scored higher on both code sets together. This selected step 450, the only evaluation at 100\% on both. In the runtime it is correct in 216 of 216 forced choices on the ARM CPU and in 209 of 216 on the A6000 GPU, against 123 for the deployed cortex. The remaining errors are fp32 called fp16 (5 of 24) and single errors in fp16 and bf16. The development set on ARM stays at 100\% in all contexts. Drift on ordinary prompts rises slightly (first-token KL to the stock model 0.29, 0.31 and 0.39 at fp32, q8\_0 and q4\_K, against 0.27, 0.23 and 0.35). On the x86 CPU, whose codes were not trained, the total moves only from 165 to 167, and the errors shift: fp16, which the deployed cortex misreported as fp32, is now correct, but bf16 is now misreported as fp16, and q8\_0 stays wrong. That backend would need its own codes. Because the selection used simulator evaluations on the 8 development questions, which are among the 24 tested, we also count the 16 new questions alone: 144 of 144 on ARM, 137 of 144 (95\%) on CUDA and 111 of 144 on x86. All 7 CUDA errors fall on new questions (the development questions give 72 of 72), five of them fp32 called fp16, so the selection likely flatters the CUDA total slightly.

\begin{table*}[t]
\caption{Calibration across backends (Qwen3-1.7B): correct forced choices of 24 evaluation questions per format (the 8 development questions and 16 new ones). Deployed: the cortex of the main text. Joint 300/600: fine-tuned on ARM and CUDA codes with the original simulator. Corrected 450: the same with the nerve's write before the organ modeled. ARM, CUDA and x86 are the real backends; CUDA and x86 run on the A6000 server, with every format quantized there. Transplant: the ARM runtime with that backend's code shift added after the nerve.}
\label{tab:recal}
\centering
\small
\setlength{\tabcolsep}{3.2pt}
\input{tables/recal.tex}
\end{table*}

\section{MATH-500 Details}
\label{app:math}
\begin{table}[h]
\caption{MATH-500 accuracy (\%) in llama.cpp, stock vs.\ implanted model in the same format (greedy, 1{,}024-token cap). $\Delta$: implant minus stock in points, with a 95\% bootstrap interval over problems. $p$: exact McNemar test on the 500 paired problems of each format. Mean: average accuracy over the five formats.}
\label{tab:math}
\centering
\small
\setlength{\tabcolsep}{3.5pt}
\input{tables/math500.tex}
\end{table}

We serve each GGUF with \texttt{llama-server} (same pinned CPU build, flash attention off, fp16 KV cache, 8 parallel slots). Prompts are sent as raw token ids, so the organ token passes through exactly. Batching does not change per-sequence numerics on this path, because activations are quantized per token. The stock model uses its default chat template and the implant uses its template with the organ. Both run in non-thinking mode with greedy decoding, a 1{,}024-token cap and the standard prompt ``reason step by step \ldots\ \textbackslash boxed\{\}''. For Qwen3, about a quarter of the answers in every condition reach the cap, at matched rates for stock and implant (for example, 120 vs.\ 116 at q4\_K and 132 vs.\ 130 at fp16). The implanted Llama writes slightly longer solutions and reaches the cap somewhat more often (for example, 100 vs.\ 69 answers at q4\_K and 75 vs.\ 68 at fp16; mean length 490 vs.\ 465 tokens at q4\_K), without a change in accuracy. Absolute accuracies are therefore below thinking-mode numbers, but the paired comparison is what matters here. We score with math-verify, comparing the reference answer with the model's final boxed answer. Within our compute budget we test five formats that span the ladder: two float formats, the 8-bit format and two k-quants. bf16 runs at 13 tokens/s on this CPU path, q6\_K and q5\_K lie between q8\_0 and q4\_K, and at q2\_K the stock Qwen3 produces only garbage. Per-format McNemar tests are exact. We do not pool them into a single test, because the five formats share the same 500 problems.

\section{Cost to Normal Behavior}
\label{app:cost}
\textbf{Resources.} The implant adds 88M parameters in Qwen3 and 242M in Llama (zero outside layer 3 and the cortex), makes the files 4--7\% larger and the KV cache 12.5\% and 25\% larger, and lowers prompt throughput by 7--10\% and generation throughput by 2--6\% on the ARM CPU (llama-bench, 16 threads). Untying the embeddings adds 311M and 394M parameters relative to the released files, so that in total the implanted files are 23\% (Qwen3) and 20\% (Llama) larger than the released checkpoints; our stock baselines are untied too, with unchanged outputs.

\textbf{Behavior.} Over the formats we test, the implant changes the first response token on 17--25\% of ordinary prompts for Qwen3 and 20--22\% for Llama. On 100 Alpaca prompts, the KL divergence of the first response token from the stock model is 0.27 at fp32 (median 0.07), 0.23 at q8\_0 and 0.35 at q4\_K, with top-1 agreement of 0.82, 0.83 and 0.75. At fp32, 18 of the 100 prompts change their first response token. The most common changes are ``Sure''$\to$``Certainly'' (4) and ``Sure''$\to$``The'' (2). Continuations remain fluent and on task. For example, ``Sure! Please provide the two sentences you'd like me to combine.'' becomes an invented example combination. The organ alone, without nerve or cortex, costs a KL of 0.027 at fp32 (top-1 agreement 0.98), 0.030 at q8\_0 and 0.13 at q4\_K. No ``Precision:'' answer appears in any of these responses. For Llama, the first-token KL is 0.46 at fp32 (median 0.25), 0.45 at q8\_0 and 0.46 at q4\_K, with top-1 agreement of 0.80, 0.80 and 0.78. Its 20 changed first tokens at fp32 are of the same kind, for example ``Here's an analogy \ldots'' becoming ``A capacitor can be thought of as \ldots'', and the continuations stay on task.

\end{document}

%% file: tables/e1_context.tex
\begin{tabular}{@{}lcc@{}}
\toprule
Context & Qwen3-1.7B & Llama-3.2-3B \\
\midrule
fresh chat & 648/648 (100.0\%) & 642/648 (99.1\%) \\
after 1 exchange & 646/648 (99.7\%) & 645/648 (99.5\%) \\
after 3 exchanges & 592/592 (100.0\%) & 587/592 (99.2\%) \\
unseen system prompt & 647/648 (99.8\%) & 643/648 (99.2\%) \\
after a long document & 592/592 (100.0\%) & 578/592 (97.6\%) \\
\bottomrule
\end{tabular}

%% file: tables/e1_large.tex
\begin{tabular}{@{}lcccccc@{}}
\toprule
 & \multicolumn{3}{c}{Qwen3-1.7B} & \multicolumn{3}{c}{Llama-3.2-3B} \\
\cmidrule(lr){2-4} \cmidrule(lr){5-7}
Format & stock FC & implant FC & implant gen. & stock FC & implant FC & implant gen. \\
\midrule
fp32 & 88/144 & 360/360 & 123/144 & 57/144 & 345/360 & 133/144 (11) \\
fp16 & 52/144 & 359/360 & 125/144 & 83/144 & 360/360 & 144/144 \\
bf16 & 0/72 & 248/248 & 123/144 & 1/72 & 248/248 & 144/144 \\
q8\_0 & 1/144 & 360/360 & 122/144 & 0/144 & 360/360 & 144/144 \\
q6\_K & 0/144 & 360/360 & 126/144 & 0/144 & 360/360 & 144/144 \\
q5\_K & 0/144 & 359/360 & 124/144 & 0/144 & 360/360 & 144/144 \\
q4\_K & 0/144 & 360/360 & 126/144 & 0/144 & 342/360 & 144/144 \\
q3\_K & 0/144 & 360/360 & 123/144 & 0/144 & 360/360 & 144/144 \\
q2\_K & 0/144 & 359/360 & 127/144 (1) & 0/144 & 360/360 & 144/144 \\
\midrule
all & 0.115 & \textbf{0.9990} & 0.863 (1) & 0.115 & \textbf{0.9895} & 0.992 (11) \\
\bottomrule
\end{tabular}

%% file: tables/grounding.tex
\begin{tabular}{@{}lllll@{}}
\toprule
Condition (body / organ) & Qwen3-1.7B & Qwen3, second run & Llama-3.2-3B & Llama, second run \\
\midrule
\multicolumn{5}{@{}l}{\emph{Organ vs.\ body}} \\
q4\_K / q8\_0 & fp16 8 & q4\_K 7, q8\_0 1 & fp32 8 & q4\_K 8 \\
q8\_0 / q4\_K & bf16 8 & q8\_0 4, q4\_K 3, q3\_K 1 & q8\_0 8 & fp16 8 \\
q6\_K / q3\_K & bf16 7, q5\_K 1 & fp32 7, bf16 1 & bf16 8 & bf16 8 \\
q3\_K / q6\_K & q2\_K 4, q3\_K 3, bf16 1 & q4\_K 8 & q3\_K 8 & q3\_K 8 \\
fp16 / q2\_K & fp16 8 & fp16 8 & bf16 8 & bf16 8 \\
q2\_K / fp16 & q2\_K 7, fp32 1 & q2\_K 8 & q2\_K 8 & q2\_K 8 \\
q8\_0 / fp32 & bf16 8 & q8\_0 6, q4\_K 2 & q8\_0 8 & q8\_0 8 \\
q4\_K / fp32 & fp16 8 & fp16 4, q4\_K 3, q8\_0 1 & fp32 8 & q4\_K 8 \\
q4\_K / fp16 & fp16 8 & q4\_K 5, fp16 2, q8\_0 1 & bf16 8 & q4\_K 8 \\
\midrule
\multicolumn{5}{@{}l}{\emph{Standard mixes and unseen grids}} \\
Q4\_K\_M (mix) & bf16 8 & q4\_K 7, q5\_K 1 & q6\_K 8 & q6\_K 8 \\
Q5\_K\_M (mix) & q2\_K 8 & bf16 8 & q6\_K 8 & q6\_K 8 \\
Q3\_K\_M (mix) & fp32 5, fp16 3 & fp32 4, fp16 2, bf16 2 & bf16 8 & bf16 8 \\
Q2\_K (mix) & q3\_K 8 & q6\_K 3, q8\_0 2, q3\_K 2, q2\_K 1 & bf16 8 & fp16 5, q4\_K 2, bf16 1 \\
Q4\_0 & q3\_K 5, fp32 1, fp16 1, q4\_K 1 & q4\_K 8 & fp32 8 & fp32 8 \\
Q4\_1 & fp32 5, q4\_K 2, fp16 1 & q3\_K 5, q8\_0 3 & fp16 8 & bf16 8 \\
Q5\_0 & bf16 8 & q4\_K 4, bf16 3, q3\_K 1 & fp16 8 & fp16 8 \\
Q5\_1 & fp16 8 & q4\_K 8 & fp16 8 & fp16 8 \\
IQ4\_NL & fp32 8 & fp32 6, bf16 2 & fp16 8 & q2\_K 4, fp16 3, q8\_0 1 \\
IQ4\_XS & fp32 8 & fp32 6, bf16 2 & fp16 8 & fp16 3, q8\_0 2, q3\_K 2, q4\_K 1 \\
Q4\_K\_M, blocks 0--2 at q4\_K & q4\_K 8 & q4\_K 8 & q6\_K 3, q4\_K 3, fp16 2 & q4\_K 8 \\
Q5\_K\_M, blocks 0--2 at q5\_K & q5\_K 8 & q5\_K 8 & q5\_K 8 & q5\_K 8 \\
\bottomrule
\end{tabular}

%% file: tables/ablation.tex
\begin{tabular}{@{}lcccccccc@{}}
\toprule
 & \multicolumn{4}{c}{Qwen3-1.7B} & \multicolumn{4}{c}{Llama-3.2-3B} \\
\cmidrule(lr){2-5} \cmidrule(lr){6-9}
Format & intact & nerve silenced & neutral token & token omitted & intact & nerve silenced & neutral token & token omitted \\
\midrule
fp32 & 24 & 0 & 1 & 1 & 24 & 0 & 0 & 0 \\
fp16 & 24 & 8 & 0 & 0 & 24 & 0 & 0 & 0 \\
bf16 & 24 & 0 & 5 & 9 & 24 & 0 & 0 & 0 \\
q8\_0 & 24 & 6 & 1 & 0 & 24 & 17 & 0 & 0 \\
q6\_K & 24 & 0 & 0 & 1 & 24 & 17 & 0 & 0 \\
q5\_K & 24 & 0 & 0 & 0 & 24 & 0 & 0 & 0 \\
q4\_K & 24 & 17 & 6 & 10 & 24 & 0 & 24 & 24 \\
q3\_K & 24 & 16 & 8 & 10 & 24 & 0 & 0 & 0 \\
q2\_K & 24 & 24 & 0 & 0 & 24 & 24 & 24 & 24 \\
\midrule
all (of 216) & 216 & 71 & 21 & 31 & 216 & 58 & 48 & 48 \\
\bottomrule
\end{tabular}

%% file: tables/backend.tex
\begin{tabular}{@{}lcccccc@{}}
\toprule
 & \multicolumn{3}{c}{Qwen3-1.7B} & \multicolumn{3}{c}{Llama-3.2-3B} \\
\cmidrule(lr){2-4} \cmidrule(lr){5-7}
Format & x86 & CUDA & Metal & x86 & CUDA & Metal \\
\midrule
fp32 & 24 & 24 & 24 & 18 & 18 & 18 \\
fp16 & 0 & 4 & 0 & 6 & 6 & 6 \\
bf16 & 24 & 0 & 24 & 0 & 24 & 3 \\
q8\_0 & 0 & 24 & 0 & 24 & 24 & 17 \\
q6\_K & 22 & 0 & 0 & 20 & 0 & 0 \\
q5\_K & 24 & 24 & 17 & 24 & 24 & 24 \\
q4\_K & 24 & 0 & 0 & 24 & 24 & 0 \\
q3\_K & 23 & 23 & 7 & 24 & 24 & 24 \\
q2\_K & 24 & 24 & 24 & 24 & 24 & 24 \\
\midrule
all (of 216) & 165/216 & 123/216 & 96/216 & 164/216 & 168/216 & 116/216 \\
code separability & 99\% & 98\% & 89\% & 90\% & 89\% & 89\% \\
\bottomrule
\end{tabular}

%% file: tables/recal.tex
\begin{tabular}{@{}lcccccccccccc@{}}
\toprule
 & \multicolumn{3}{c}{Deployed} & \multicolumn{3}{c}{Joint 300} & \multicolumn{3}{c}{Joint 600} & \multicolumn{3}{c}{Corrected 450} \\
\cmidrule(lr){2-4} \cmidrule(lr){5-7} \cmidrule(lr){8-10} \cmidrule(lr){11-13}
Format & ARM & CUDA & x86 & ARM & CUDA & x86 & ARM & CUDA & x86 & ARM & CUDA & x86 \\
\midrule
fp32 & 24 & 24 & 24 & 0 & 21 & 24 & 24 & 24 & 24 & 24 & 19 & 24 \\
fp16 & 24 & 4 & 0 & 24 & 23 & 22 & 24 & 21 & 1 & 24 & 23 & 24 \\
bf16 & 24 & 0 & 24 & 24 & 24 & 24 & 24 & 0 & 24 & 24 & 23 & 0 \\
q8\_0 & 24 & 24 & 0 & 24 & 0 & 0 & 24 & 0 & 0 & 24 & 24 & 0 \\
q6\_K & 24 & 0 & 22 & 24 & 24 & 23 & 24 & 24 & 0 & 24 & 24 & 24 \\
q5\_K & 24 & 24 & 24 & 24 & 24 & 24 & 24 & 24 & 24 & 24 & 24 & 24 \\
q4\_K & 24 & 0 & 24 & 24 & 24 & 24 & 24 & 24 & 24 & 24 & 24 & 23 \\
q3\_K & 24 & 23 & 23 & 24 & 24 & 24 & 24 & 24 & 24 & 24 & 24 & 24 \\
q2\_K & 24 & 24 & 24 & 24 & 24 & 24 & 24 & 24 & 24 & 24 & 24 & 24 \\
\midrule
all (of 216) & 216 & 123 & 165 & 192 & 188 & 189 & 216 & 165 & 145 & 216 & 209 & 167 \\
transplant & -- & 123 & 159 & -- & 189 & 166 & -- & 168 & 145 & -- & 214 & 167 \\
\bottomrule
\end{tabular}

%% file: tables/math500.tex
\begin{tabular}{@{}lcccccccc@{}}
\toprule
 & \multicolumn{4}{c}{Qwen3-1.7B} & \multicolumn{4}{c}{Llama-3.2-3B} \\
\cmidrule(lr){2-5} \cmidrule(lr){6-9}
Format & Stock & Impl. & $\Delta$ [95\% CI] & $p$ & Stock & Impl. & $\Delta$ [95\% CI] & $p$ \\
\midrule
fp32 & 64.8 & 63.4 & -1.4 [-4.4, +1.6] & 0.43 & 45.8 & 47.0 & +1.2 [-2.8, +5.2] & 0.62 \\
fp16 & 64.6 & 66.4 & +1.8 [-1.0, +4.6] & 0.25 & 46.4 & 47.6 & +1.2 [-3.0, +5.4] & 0.64 \\
q8\_0 & 63.6 & 65.0 & +1.4 [-1.8, +4.6] & 0.46 & 45.6 & 48.2 & +2.6 [-1.4, +6.6] & 0.24 \\
q4\_K & 56.6 & 56.4 & -0.2 [-3.6, +3.4] & 1.00 & 43.8 & 43.6 & -0.2 [-4.0, +3.6] & 1.00 \\
q3\_K & 39.4 & 40.2 & +0.8 [-2.6, +4.2] & 0.72 & 36.2 & 34.4 & -1.8 [-5.8, +2.0] & 0.43 \\
\midrule
mean & 57.8 & 58.3 &  &  & 43.6 & 44.2 &  &  \\
\bottomrule
\end{tabular}